\documentclass[11pt]{asiresearch}
\usepackage[T1]{fontenc}
\usepackage{XCharter}

\usepackage{float}
\makeatletter
\patchcmd{\@maketitle}{\vskip 2em}{\vskip 0pt}{}{
  \PackageError{wwm}{Could not adjust title spacing}{Check the article title definition.}}
\patchcmd{\@maketitle}{\vskip 1em{\large \@date}}{}{}{
  \PackageError{wwm}{Could not remove the empty date row}{Check the article title definition.}}
\makeatother
\usetikzlibrary{calc, shadows, shapes.geometric, decorations.markings, backgrounds, fadings, positioning}

\definecolor{apBlue}{HTML}{007AFF}
\definecolor{apPurple}{HTML}{AF52DE}
\definecolor{apPink}{HTML}{FF2D55}
\definecolor{apRed}{HTML}{FF3B30}
\definecolor{apOrange}{HTML}{FF9500}
\definecolor{apYellow}{HTML}{FFCC00}
\definecolor{apGreen}{HTML}{28CD41}
\definecolor{apTeal}{HTML}{59ADC4}
\definecolor{apGray}{HTML}{8E8E93}

\definecolor{hsrDark}{HTML}{0F1018} % Deep space
\definecolor{hsrGold}{HTML}{D7B368} % UI Gold accents
\definecolor{hsrGlow}{HTML}{5D6D7E} % Faint nebula

\usepackage{algorithm}
\usepackage{algpseudocode}% Provides the algorithmic commands like \Require, \State, etc.

\ifdefined\final
\usepackage[disable]{todonotes}
\else
\usepackage[textsize=tiny]{todonotes}
\fi

\title{\huge \bfseries Web World Models}
\author{\textbf{Jichen Feng}\thanks{Equal contribution;~~$^\dagger$Corresponding authors.}~$^{1,3}$~~~~\textbf{Yifan Zhang}\footnotemark[1]~$^{1}$~~~~\textbf{Chenggong Zhang}\footnotemark[1]~$^{2}$~~~~\textbf{Yifu Lu}\footnotemark[1]~$^1$\\[0.5mm]
\textbf{Shilong Liu}$^1$$^{\dagger}$~~~~\textbf{Mengdi Wang}$^1$$^{\dagger}$\\[1.5mm]
$^1$Princeton University~~~~$^2$University of California, Los Angeles~~~~\\[0.5mm]$^3$University of Pennsylvania}
\date{}

\begin{document}
\maketitle

%%%%%%%%%%%%%%%%%%%%%%%%%%%%%%
%%% Abstract
\begin{abstract}
Language agents increasingly require persistent worlds in which they can act, remember, and learn. Existing approaches sit at two extremes: conventional web frameworks provide reliable but fixed contexts backed by databases, while fully generative world models aim for unlimited environments, but the world is constructed primarily through generation, making it harder to maintain a fixed, deterministic global framework, reducing controllability.
In this work, we introduce the \textbf{Web World Model} (WWM), a middle ground where world state and ``physics'' are implemented in ordinary web code to ensure logical consistency, while large language models generate context, narratives, and high-level decisions on top of this structured latent state. We build a suite of WWMs on a realistic web stack, including an infinite travel atlas grounded in real geography, fictional galaxy explorers, web-scale encyclopedic and narrative worlds, and simulation- and game-like environments. Across these systems, we identify practical design principles for WWMs: separating code-defined rules from model-driven imagination, representing latent state as typed web interfaces, and utilizing deterministic generation to achieve unlimited but structured exploration. Our results suggest that web stacks themselves can serve as a scalable substrate for world models, enabling controllable yet open-ended environments.
\end{abstract}

\projectpage{https://princeton-ai2-lab.github.io/Web-World-Models/}

% Keep the overview and its caption together on the first page.
\begin{figure}[H]
\centering
\includegraphics[width=0.7\linewidth]{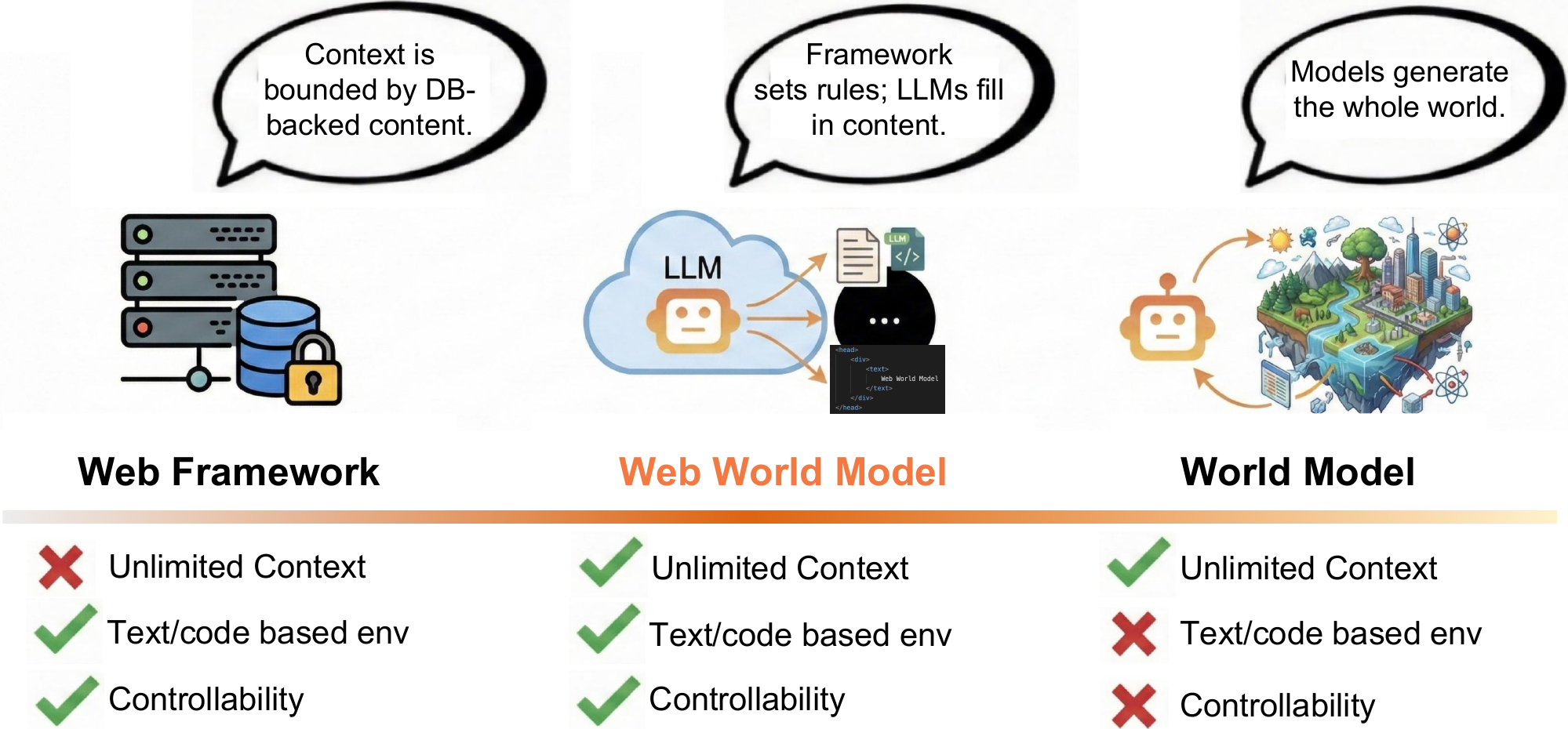}
\caption{\textbf{Left}: Traditional Web Frameworks fix context in databases, limiting scalability. \textbf{Center}: The Web World Model (Ours) decouples logic from content, generating unlimited context via LLMs upon a code-based physics layer without heavy data storage. \textbf{Right}: Fully generative world models can produce unlimited context and rich video/3D content, but when the world is constructed primarily through generation, it is harder to maintain a fixed, deterministic global framework, reducing controllability.}
\label{fig:overview}
\end{figure}
\clearpage

%%%%%%%%%%%%%%%%%%%%%%%%%%%%%%
%%% Introduction

\section{Introduction}

Modern language agents increasingly need persistent environments in which they can act, remember, and grow. Today, most practical systems sit at one of two extremes (Figure~\ref{fig:overview}).
On one side, conventional~\emph{web frameworks} operate with a fixed context: application state is stored in databases and exposed through hand-crafted endpoints. This design offers reliability, robust engineering tooling, and clear security boundaries, but the world agents can inhabit is ultimately bounded by the schema developers anticipated in advance. On the other side, \emph{general world models} attempt to generate environments directly in the latent space of a model, in principle supporting unlimited context and arbitrary environment types.
However, these fully generative worlds are difficult to control, hard to debug, and costly to scale, and they often lack the structural guarantees needed for long-running applications.
As a result, there is a missing middle ground between fixed-context web applications and unconstrained world models.

We propose to fill this gap with the notion of a \textbf{Web World Model} (WWM).
A Web World Model is a world whose state and ``physics'' are defined by ordinary web code (e.g., TypeScript modules, HTTP handlers, and database schemas), while large language models generate context and narratives on top of this structured latent state.
In this view, code specifies what kinds of entities exist, how they interact, and which actions are possible; the model is invoked to enrich these entities with descriptions, stories, or task-specific reasoning.
WWMs thus inherit the controllability, observability, and tooling of web frameworks, yet they can procedurally expand to an effectively unlimited state space by using language models to synthesize new content on demand.
Compared with fixed web systems, WWMs are not bound to a small, static context; compared with fully generative world models, they offer a programmable substrate that can be tested, versioned, and deployed using standard web infrastructure.

\begin{figure}[ht!]
    \centering
    \includegraphics[width=0.8\linewidth]{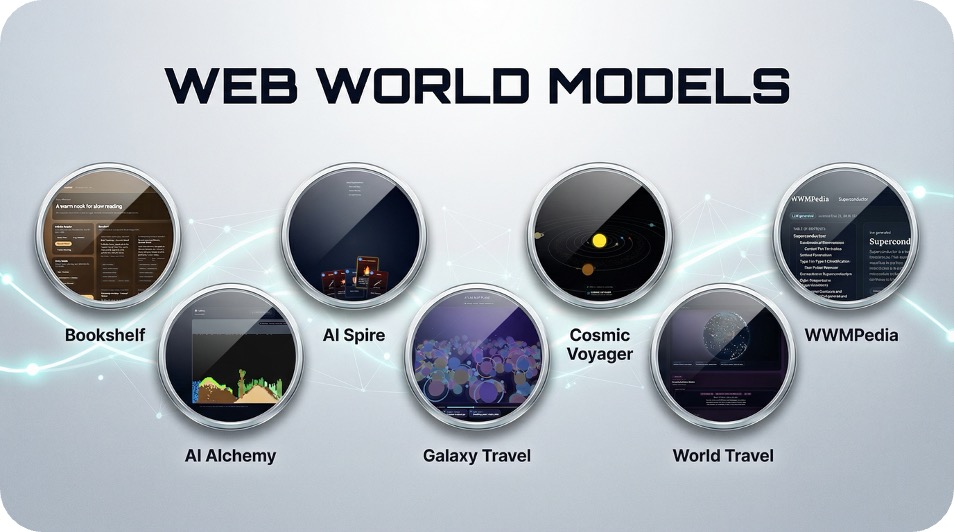}
    \caption{\textbf{Illustrations of a series of Web World Models introduced in this work.} Details are presented in Section~\ref{sec:examples}.}
    \label{fig:Teaser1}
\end{figure}

To make this idea concrete, we implement a series of Web World Models on a unified
web technology stack.
These systems span a wide range of domains.
An \emph{Infinite Travel Atlas} turns the real Earth into an explorable atlas, where any coordinate can be expanded into rich places, routes, and stories.
Fictional-universe demos such as \emph{Galaxy Travel} generate procedural galaxies whose large-scale structure is governed by code, while language models provide missions, characters, and educational content.
Other demos wrap the open web as a browsable environment, or reinterpret long-form books as navigable narrative worlds.
We also build simulation- and game-like environments, including alchemy-style combinatorial systems and card-based roguelikes, that treat the language model as a rule and content generator while keeping core transitions deterministic. Together, these demos illustrate that the Web World Model abstraction is not tied to a particular task or genre: it can host worlds that are real or fictional, knowledge-centric or interaction-driven, single-user or multi-agent.

Across these examples, we distill four core design principles for building Web World Models (Section~\ref{sec:design}). First, Separation of Concerns: core rules and state transitions (Physics) must be distinct from creative generation (Imagination). Second, Typed Interfaces: latent world states should be represented as explicit, typed web interfaces (JSON schemas) rather than opaque embeddings. Third, Infinite Worlds via Deterministic Generation: procedural expansions must respect a fixed schema to allow the world to grow without exploding the action space. Finally, Graceful Degradation: by combining procedural generators and templates, worlds must remain usable even when model calls are slow or unavailable.

In summary, this paper explores how web stacks themselves can host world models by treating code as the substrate of physics and language models as bounded imagination engines. Our main contributions are:
\begin{itemize}
\item We formulate the concept of a \textbf{Web World Model} and position it between
fixed-context web frameworks and fully generative world models, clarifying the
design space along the axes of context capacity, controllability, and environment type.
\item We implement a suite of Web World Models on a realistic web technology stack,
spanning real-world atlases, fictional galaxies, web-scale encyclopedias,
interactive narratives, simulations, and games, demonstrating the generality
of the abstraction.
\item We derive a set of practical design insights for building WWMs, Separation of Concerns, Typed Interfaces, Deterministic Generation, and Graceful Degradation, and discuss how these insights guide future LLM-based environment and agent design.
\end{itemize}

\section{Design Principles of Web World Models}
\label{sec:design}

Reliable environmental modeling requires a synergy between deterministic logic and probabilistic generation. Purely generative models are prone to hallucination and state inconsistency, while traditional hard-coded environments lack semantic flexibility and open-endedness. The \textbf{Web World Model} (WWM) bridges this gap via a hybrid architecture. We delineate four core principles that render this framework practical for scalable deployment.

\begin{figure}[ht]
    \centering
    \begin{tikzpicture}[
        node distance=1.5cm,
        >=stealth,
        font=\small\sffamily,
        block/.style={rectangle, draw=gray!60, fill=white, text width=2.5cm, align=center, rounded corners, minimum height=1.2cm, thick},
        arrow/.style={->, thick, color=gray!80},
        codeblock/.style={block, fill=blue!5, draw=blue!40},
        aiblock/.style={block, fill=purple!5, draw=purple!40},
        label/.style={text=gray!70, font=\footnotesize}
    ]
        % Nodes
        \node[block, text width=1.5cm] (input) {User Action\\($a_t$)};
        
        % The Physics Box
        \node[codeblock, right of=input, node distance=4.5cm] (physics) {\textbf{Physics ($S^{\phi}$)}\\Deterministic Code\\Inventory, Map, Logic};
        
        % The Imagination Box
        \node[aiblock, right of=physics, node distance=5cm] (imagination) {\textbf{Imagination ($S^{\psi}$)}\\Stochastic LLM\\Description, Dialogue, Vibe};
        
        % Output
        \node[block, right of=imagination, node distance=4.5cm, text width=1.5cm] (output) {Rendered\\World};

        % Flow
        \draw[arrow] (input) -- node[above, font=\scriptsize] {Triggers Update} (physics);
        \draw[arrow] (physics) -- node[above, font=\scriptsize] {Context ($S^{\phi}_{t+1}$)} (imagination);
        \draw[arrow] (imagination) -- node[above, font=\scriptsize] {Narrative ($S^{\psi}_{t+1}$)} (output);
        
        % Feedback loop representation
        \draw[arrow, dashed, bend left=45] (output.south) to node[below, font=\scriptsize] {Loop: New State} (input.south);

        % Annotations
        \node[below of=physics, node distance=1.5cm, text=blue!60, font=\scriptsize] {Strict Logic};
        \node[below of=imagination, node distance=1.5cm, text=purple!60, font=\scriptsize] {Creative Layer};

    \end{tikzpicture}
    \caption{The Web World Model Architecture: A separation between the deterministic Code Layer (Physics) and the stochastic AI Layer (Imagination).}
    \label{fig:wwm-architecture}
\end{figure}
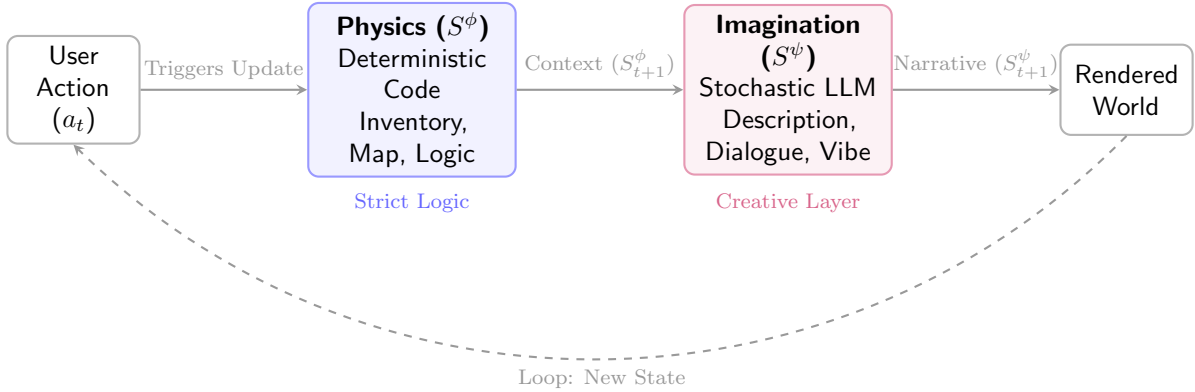

\subsection{Separation of Concerns: Physics vs. Imagination}

Analogous to the distinction between the physics engine and the rendering pipeline in video games, a WWM decomposes the world state $S_t$ into two orthogonal components: $S_t = (S^{\phi}_t, S^{\psi}_t)$. The \textbf{Physics Layer} ($S^{\phi}$) is defined strictly by deterministic code. It maintains invariant state data: inventories, coordinates, resource caps, and enforces logical consistency (e.g., preventing transit through locked doors or spending deficit funds). The \textbf{Imagination Layer} ($S^{\psi}$) is stochastic and model-defined. It generates high-dimensional perceptual content, such as environmental descriptions, NPC dialogue, and aesthetic style.

State transitions occur in a strict order. First, the code computes the logical outcome:
\[
S^{\phi}_{t+1} = f_{\texttt{code}}(S^{\phi}_t, a_t).
\]
Subsequently, the large language model $\pi_{\theta}$ synthesizes the perceptual layer conditioned on this updated symbolic state:
\[
S^{\psi}_{t+1} \sim \pi_{\theta}(\cdot \mid S^{\phi}_{t+1}).
\]
This decomposition maintains logical consistency of the world through code, while allowing the AI to generate rich and diverse scenes and text.

\subsection{Typed Interfaces as the Common Language}

In conventional deep learning, latent states are typically represented as opaque, high-dimensional vectors. In a WWM, we replace this black-box representation with a \textbf{Typed Interface}. We define strict schemas (e.g., \texttt{interface Planet \{biome: string; hazard: string;\}}) that serve as a binding contract between the code and the model.

Rather than predicting pixels or embeddings, the LLM predicts valid JSON objects conforming to these type definitions. This transforms the latent space into a transparent, debuggable data structure. It ensures that \textit{imagined} content remains structurally compatible with the engine's logic—if the model generates a new item, it must populate the specific fields (e.g., \texttt{weight}, \texttt{cost}) required by the physics engine. The typed interface thus acts as a syntactic filter, eliminating structural hallucinations and preventing model outputs from violating application logic.

\subsection{Infinite Worlds via Deterministic Hashing}

We cannot store an infinite universe in a database. Instead, we generate it ``Just-In-Time'' using procedural generation principles. 

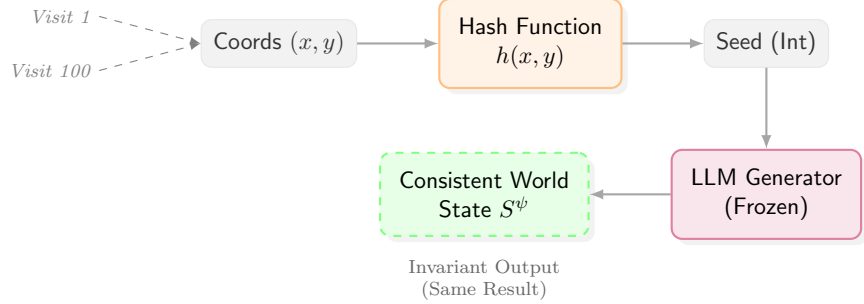
\begin{figure}[ht]
    \centering
    \begin{tikzpicture}[
        % Global settings
        scale=0.9, transform shape,
        font=\sffamily\small,
        >=latex, % standard LaTeX arrow head
        node distance=1.2cm and 1.2cm, % Vertical and Horizontal spacing
        % Styles
        process/.style={
            rectangle, draw=black!60, fill=white, thick, 
            rounded corners, minimum height=3em, inner sep=8pt, 
            align=center, drop shadow={opacity=0.1}
        },
        data/.style={
            rectangle, draw=black!10, fill=gray!10, 
            text=gray!40!black, inner sep=5pt, rounded corners
        },
        line/.style={
            ->, thick, gray!70, rounded corners=5pt
        }
    ]
    
    % ,  Nodes , 
    
    % 1. Input
    \node[data] (coords) {Coords $(x,y)$};
    
    % 2. Hash Function (Right of Input)
    \node[process, fill=orange!10, draw=orange!50, right=of coords] (hash) {Hash Function\\$h(x,y)$};
    
    % 3. Seed (Right of Hash)
    \node[data, right=of hash] (seed) {Seed (Int)};
    
    % 4. LLM (Below Seed)
    \node[process, fill=purple!10, draw=purple!50, below=of seed] (llm) {LLM Generator\\(Frozen)};
    
    % 5. Output (Left of LLM)
    \node[process, fill=green!10, draw=green!50, dashed, left=of llm] (world) {Consistent World\\State $S^{\psi}$};

    % ,  Connections , 
    \draw[line] (coords) -- (hash);
    \draw[line] (hash) -- (seed);
    \draw[line] (seed) -- (llm);
    \draw[line] (llm) -- (world);
    
    % ,  Annotations , 
    
    % Inputs (Visit 1 / 100) - Positioning them relative to Coords
    \node[font=\scriptsize\itshape, gray, anchor=east] (v1) at ([xshift=-1.5cm, yshift=0.4cm]coords.west) {Visit 1};
    \node[font=\scriptsize\itshape, gray, anchor=east] (v100) at ([xshift=-1.5cm, yshift=-0.4cm]coords.west) {Visit 100};
    
    % Arrows converging on Coords
    \draw[dashed, gray, ->] (v1.east) -- (coords.west);
    \draw[dashed, gray, ->] (v100.east) -- (coords.west);
    
    % "Same Result" Label - Moved BELOW the green box to avoid overlap
    \node[below=0.2cm of world, font=\scriptsize, text=gray!80!black, align=center] 
        {Invariant Output\\(Same Result)};

    \end{tikzpicture}
    \caption{Deterministic Generation: The inputs (visits) converge on the coordinates, passing through the hash function to produce a frozen seed. This forces the LLM to output the same world state every time.}
    \label{fig:hashing_v2}
\end{figure}

A user arrives at a location $x$. The system skips any database lookup. We pass the coordinate through a hash function and get a seed $h(x)$. The seed fixes the LLM's sampling randomness. A player can leave a planet, come back later, and the planet stays the same. This gives \textbf{Object Permanence} with no storage cost:
\begin{equation}
S^{\psi}_{t} \equiv S^{\psi}_{t+k} \quad \text{if} \quad \text{location}(t) = \text{location}(t+k).
\end{equation}

\subsection{Graceful Degradation}

Invoking an LLM for every frame is computationally prohibitive. A WWM employs a \textbf{Fidelity Slider} to adapt to resource constraints. At \textit{High Fidelity}, the LLM generates bespoke content in real-time. Under latency constraints, the system degrades to \textit{Medium Fidelity}, retrieving cached content. In the worst-case scenario, it falls back to \textit{Base Fidelity}, where deterministic code utilizes pre-authored templates.

Because code governs the Physics ($S^{\phi}$), the application remains functional even if the Imagination ($S^{\psi}$) layer becomes unavailable. The environment may lose semantic richness, but logical continuity is preserved.

\subsection{The Technical Stack}

The modern web stack provides an ideal substrate for WWMs. TypeScript provides the type safety required for the neuro-symbolic contract; HTTP streaming allows for real-time text delivery; and serverless architecture enables the infinite scaling of procedurally generated worlds without persistent infrastructure management.

\section{Examples}
\label{sec:examples}

To demonstrate the versatility of the Web World Model framework, we developed a suite of applications spanning diverse domains, from faithful geographic simulations to open-ended fictional narratives and logic-driven games. Each example specifically instantiates the design principles from Section~\ref{sec:design}: separating physics from imagination, utilizing typed interfaces for state, and employing deterministic generation for scalability.

\subsection{Infinite Travel Atlas}

\begin{figure}[ht!]
    \centering
    \includegraphics[width=1.0\linewidth]{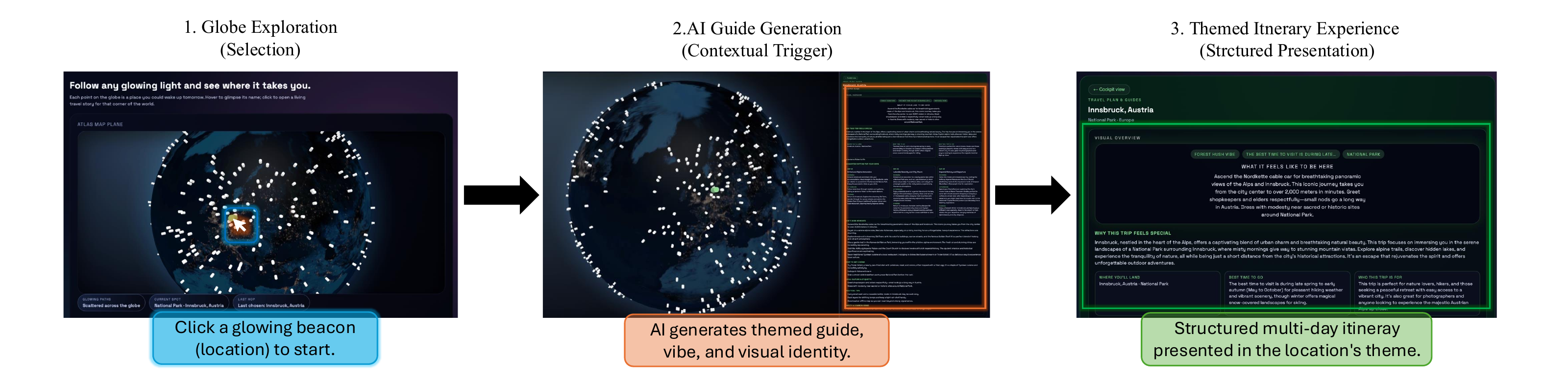} 
    \caption{\emph{Infinite Travel Atlas} interaction flow. (a) The user selects a geographic coordinate on the reactive globe. (b) This input is hashed to retrieve a deterministic seed and location metadata, grounding the request in the Physics layer ($S^{\phi}$). (c) The Agent receives this context and triggers the LLM (Imagination layer $S^{\psi}$) to generate a structured destination guide, determining visual themes and itinerary details. (d) The resulting content is rendered on the client, providing a cohesive, persistent experience without backend storage.}
    \label{fig:World-Travel}
\end{figure}

\noindent To validate the practical capabilities of a Web World Model, we created the high-fidelity interactive ecosystem-based \emph{Infinite Travel Atlas}, inspired by the famous project Google Earth. Our goal is to make the web world travel build an open exploration experience, where users have an infinite state space that can run without relying on limited stored database backends. We demonstrate how a web-based simulator can respond to the details of a traveling plan and guide the user by interacting and exploring the globe with LLM. We find that this implementation shows that the architecture can manage an infinite state space and ensure that any geographic coordinate remains accessible.

\paragraph{Environment.}
The simulation environment is a client-side TypeScript (JavaScript) application and is styled with CSS and HTML. Unlike static maps, this interface is a continuously navigated map that the user can move around everywhere. We design that the lightspot will determine the semantic context from the LLM API. The globe indicates that when the user zooms in or moves the cursor over the area of interest, additional information is rendered, thereby facilitating a more comprehensive understanding of the traveling details, which pertain to the characteristics of the area and its traveling details. We ensure that the design of every interaction is grounded in a realistic depiction of the world so that the environment can establish a connection between the global and the specific details of each node.

\paragraph{Neural-symbolic Web World Model.}
As shown in Figure~\ref{fig:World-Travel}, the system makes the Web World Model a hybrid simulator wherein the latent state comprises interpretable and executable TypeScript modules. In this architecture, code builds the semantic grounding by inferring physical attributes before selecting aesthetic themes. An LLM policy subsequently operates within this structured latent space to execute high-level creative tasks like composing visual overviews. The process of selecting a theme demonstrates creativity. First, the system computes a valid subset of themes based on deterministic code. Then, the LLM selects a specific direction to guide content generation. As these governing rules exist as executable code, changing the thematic inventory immediately reshapes the laws of the world, without the need for model retraining.

\paragraph{Agents.}
Our primary objective is to facilitate infinite exploration, allowing users to select arbitrary global coordinates while maintaining a navigable environment. We achieve this via a two-stage procedural generation strategy. First, \texttt{worldPromptService.ts} initializes the experience with query templates. Second, \texttt{proceduralBeaconService.ts} deterministically generates beacons with stable identifiers and metadata upon user interaction. This ensures destinations are not constrained by a finite database. Instead, the system synthesizes a consistent latent state JIT (Just-In-Time). The LLM consumes the beacon’s metadata as structured input to generate the destination guide. This architecture enables the agent to provide rich, creative content while symbolic code preserves geographic continuity and interface stability.

\paragraph{Demonstrations.}
Empirical observation confirms that the system maintains thematic consistency across diverse locations (Figure \ref{fig:atlas-nairobi}). Selecting a beacon near Nairobi correctly triggers a ``desert-bloom'' theme, with the LLM generating an itinerary balancing outdoor trails and history. Conversely, locations such as Honolulu (Figure \ref{fig:atlas-honolulu}) and Rio de Janeiro (Figure \ref{fig:atlas-rio}) evoke distinct thematic responses aligned with their geography while maintaining structural interface fidelity. These examples demonstrate that the WWM can transform standard web data into a controllable, infinite environment. To further test the boundaries of this abstraction, we next investigate whether this neural-symbolic design can support a purely fictional universe.

\subsection{Galaxy Travel Atlas}

\begin{figure}[ht!]
    \centering
    \includegraphics[width=1.0\linewidth]{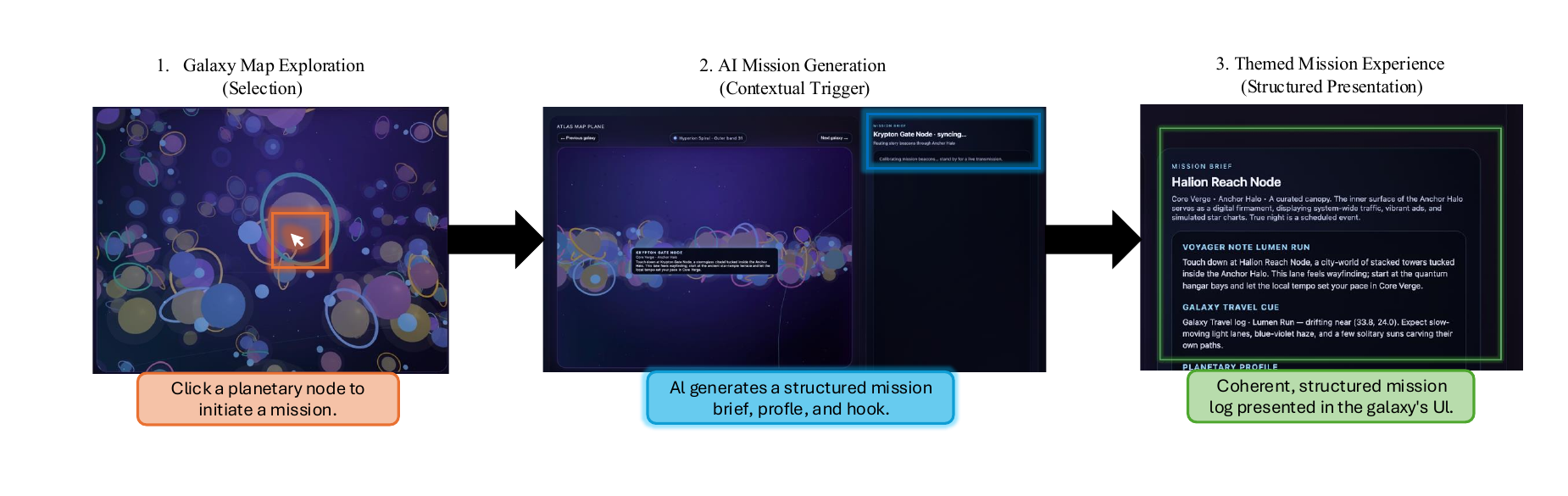}
    \caption{\emph{Galaxy Travel Atlas} system architecture. Navigation events trigger a procedural generation sequence where the coordinate hash seeds the local star system layout (Physics). The Agent retrieves this immutable structural state and queries the LLM to synthesize narrative elements, mission briefs, hazards, and lore, conforming to a strictly typed JSON schema. This ensures that while the universe is infinite, every planet remains revisited and logically consistent.}
    \label{fig: Galaxy Travel}
\end{figure}

Building upon the \emph{Infinite Travel Atlas}, we developed the \emph{Galaxy Travel Atlas}. While the Earth atlas utilizes real-world geography, this application represents a procedural sci-fi simulation where all content is synthesized. Users navigate a universe of swirling galaxies, adjusting cluster parameters and selecting planets to access mission briefs. The system leverages an LLM to generate field logs, describing terrain, sky, signals, and hazards, along with narrative hooks. This demonstration aims to validate the WWM's capacity to support a logically structured, infinite fictional universe.

\paragraph{Environment.}
The system is implemented on a TypeScript stack. The view is driven by \texttt{universe.ts}, which procedurally generates galaxy layouts, star lanes, and planet clusters. Generator parameters (e.g., planet density) are exposed to the user, ensuring dense sets of clickable worlds. A reseeding mechanism allows users to advance the generator, rendering the number of reachable galaxies effectively unbounded.

\paragraph{Neural-symbolic Web World Model.}
Following our design principles (Fig.~\ref{fig:wwm-architecture}), the \emph{Galaxy Travel Atlas} explicitly splits the world state into a deterministic \emph{physics} layer and a stochastic \emph{imagination} layer, $S_t=(S_t^{\phi},S_t^{\psi})$.
Crucially, the structural skeleton of the universe is not hallucinated; it is computed. The \emph{Physics Layer} ($S^{\phi}$) serves as the primary architect, utilizing procedural noise functions (e.g., in \texttt{universe.ts}) to dictate galaxy layouts, star lane connectivity, and planetary resource distributions. Each planet is assigned a stable identifier and a rigid set of symbolic attributes—sector labels, physical types, and risk profiles—derived purely from code.
Object permanence is achieved via hashing (Fig.~\ref{fig:hashing_v2}), ensuring that revisiting a coordinate ($x,y$) always yields the same physical state without database lookups.

The AI is relegated strictly to the \emph{Imagination Layer} ($S^{\psi}$), invoked only to texture this rigid geometry with narrative flavor text. Even then, the model is constrained by strict TypeScript interfaces (e.g., \texttt{interface Planet}); it must output valid JSON matching the code-defined biome and hazard types. If the model fails or is unreachable, the system gracefully degrades to template-based descriptions, proving that the world's existence is independent of the generative model.

\paragraph{Agents.}
The agent architecture prioritizes engineering robustness over autonomous reasoning. Agents function as stateless transformation pipelines that convert the deterministic seed and metadata into renderable JSON. By enforcing a strict schema contract via the \texttt{AgentPlugin} interface, we ensure that the generated content: mission briefs or dialogue, is structurally indistinguishable from hard-coded data.

This design allows the backend to treat the LLM as just another microservice. When a provider key is configured, the pipeline hydrates the world with bespoke text; otherwise, it falls back to static generators. Persistence is handled via file-backed caches keyed by the procedural seed, minimizing inference costs. Consequently, the ``intelligence'' of the agent is contained within a safe, verifiable sandbox, preventing the model from altering the fundamental rules or geometry of the galaxy.

\paragraph{Demonstrations.}
We present a visual traversal of the generated universe in Figures~\ref{fig:sci-fi-demo1-landing} through \ref{fig:sci-fi-demo10-haloanchor}. These snapshots demonstrate the system's ability to maintain structural invariants while generating diverse semantic content. For instance, selecting the \emph{Velis Minor Node} (Figure~\ref{fig:sci-fi-demo3-velis}) instantiates a ``stormglass'' biome characterized by crystalline hazards and specific signal patterns. In contrast, the \emph{Threx Drift Node} (Figure~\ref{fig:sci-fi-demo5-threx}) resolves into a scrap-yard metropolis with distinct risk profiles and narrative hooks. Further exploration reveals locations such as \emph{Yaka Outpost} (Figure~\ref{fig:sci-fi-demo6-yaka}) and the \emph{Halo Corridor Anchor} (Figure~\ref{fig:sci-fi-demo10-haloanchor}), which, despite originating from different procedural seeds, adhere to the same strictly typed interface. Collectively, these illustrations confirm that the Web World Model can support an effectively unbounded fictional state space, where the deterministic code layer enforces navigational continuity, and the language model populates the world with coherent, context-aware details.

\subsection{A Card Game Called AI Spire}
\begin{figure}[ht!]
    \centering
    \includegraphics[width=0.8\linewidth]{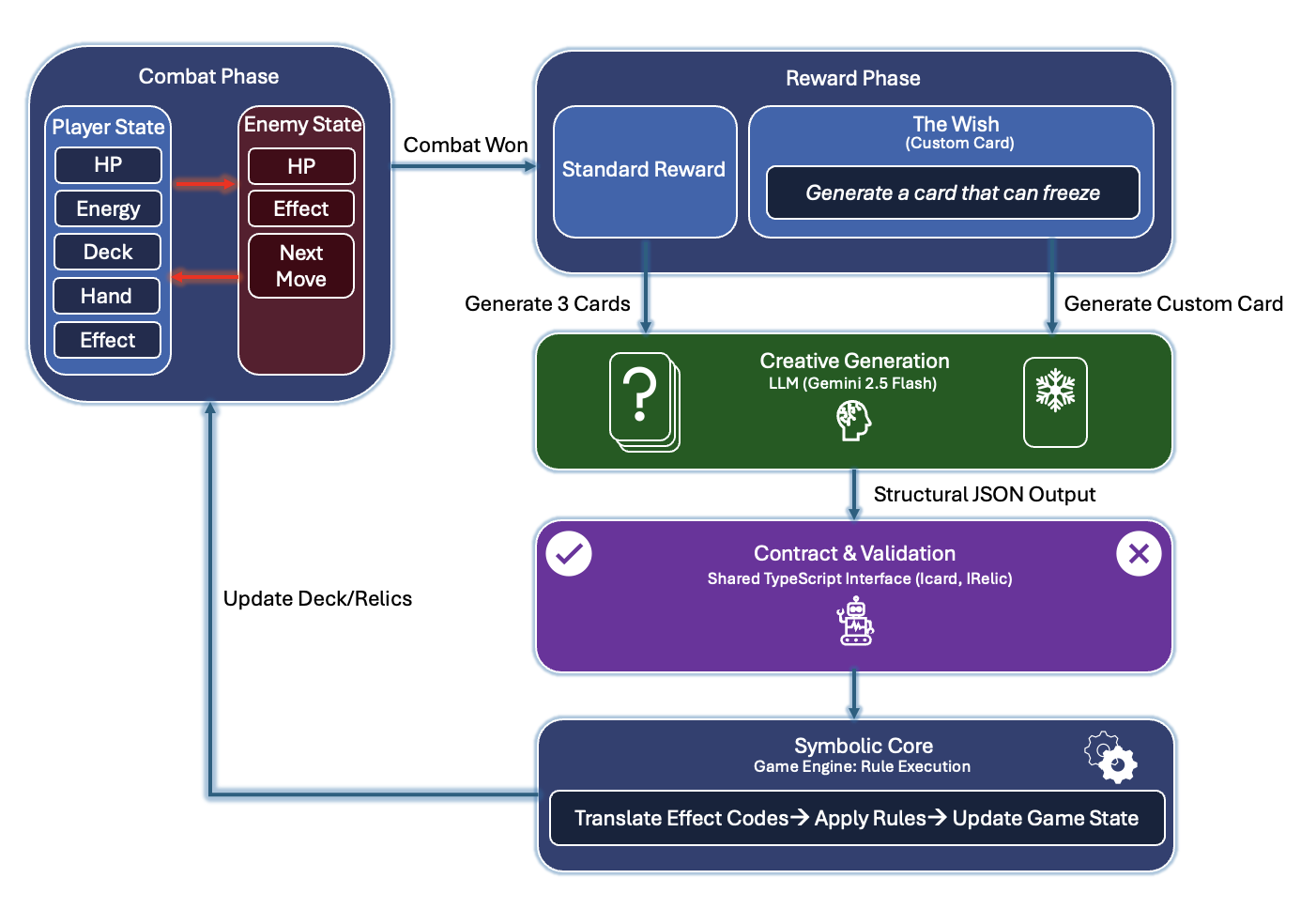}
    \caption{Neural-symbolic architecture of \emph{AI Spire}. A client-side TypeScript/React combat engine maintains player/enemy state (HP, energy, deck/hand, status effects, and enemy intent); when combat is won, the reward phase triggers either a standard reward (three generated cards) or a user-specified ``Wish'' for a custom card. In both cases, Gemini Flash acts as a constrained designer and returns a schema-structured JSON specification (name, description, and effect codes), which is checked by a contract-and-validation layer shared via TypeScript interfaces (\texttt{ICard}/\texttt{IRelic}). The symbolic core then translates effect codes into deterministic rule execution and updates the deck/relic inventory, closing the loop for the next encounter.}
    \label{fig:AI spire}
\end{figure}

\begin{figure}[ht!]
    \centering
    \includegraphics[width=0.8\linewidth]{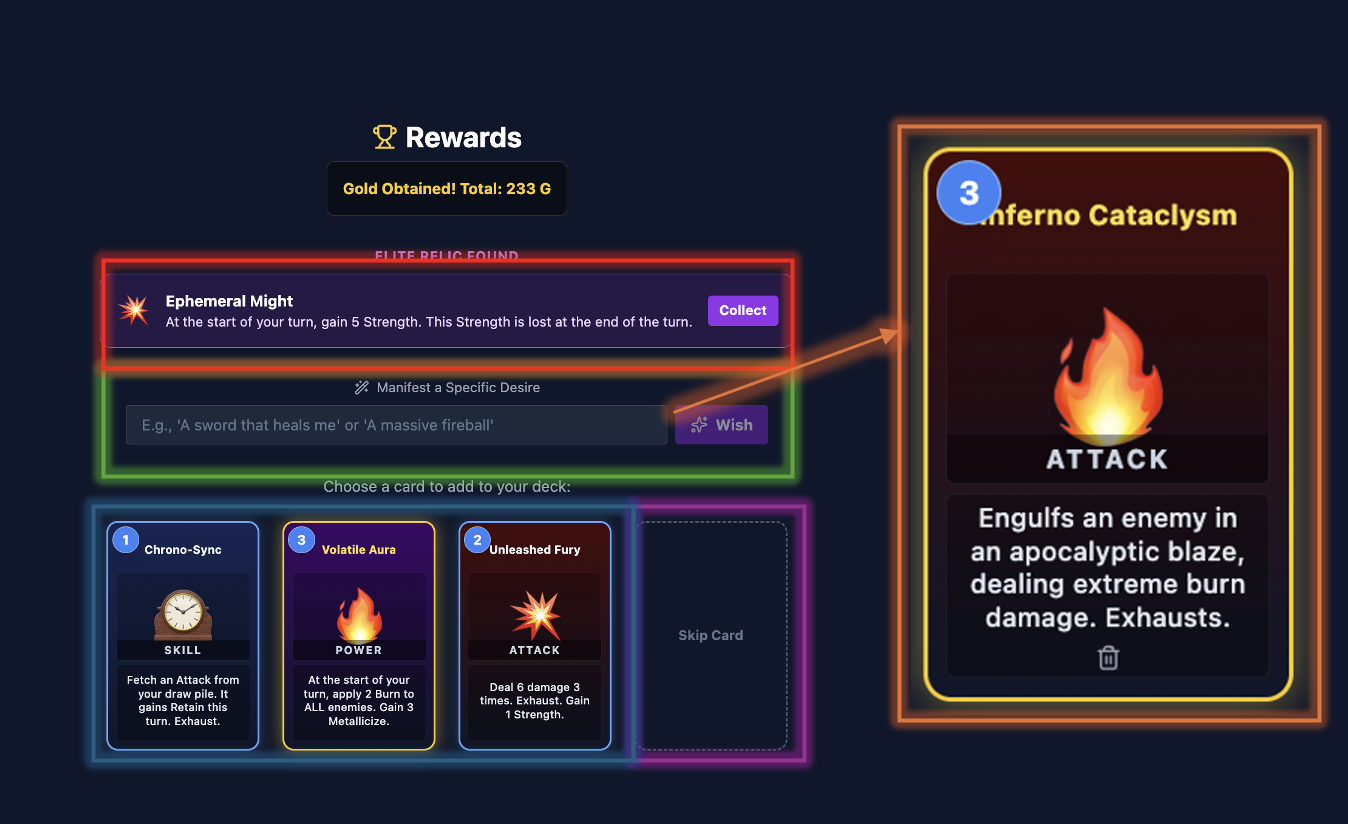}
    \caption{User interface of the reward page of \emph{AI Spire}. After winning an elite combat, an extra lyrics reward is generated by LLM (red box). Three common cards are generated by LLM and provided to the user (blue box). Besides the common rewards, users could also set a ``wish'' (green box) to customize the reward card, such as typing ``a card that could deal a huge amount of burn to the enemy.'' and then the neural-symbolic core will generate a card based on the user's request, as shown in the orange box. User could also choose to skip the reward (purple box), as in a classical \emph{Slay the Spire}-like game.}
    \label{fig:AI Spire1}
\end{figure}

\paragraph{Environment.}
The client-side of the system is TypeScript/React 19 and styled with Tailwind CSS, driven by the Google GenAI SDK. The system does not have a backend database for the reward picking (including cards and relics), instead, the reward content is generated from user's requests and must meet the expected typed interface of the renderer, e.g., the prompts and JSON schema that control the reward generation under \texttt{services/geminiService.ts} and the core engine state (player, enemy, deck/discard, and round turning) maintained by \texttt{App.tsx}. The generating and executing process shares the TypeScript (\texttt{ICard} and \texttt{IRelic}) and forms an explicit contract in order to reduce the runtime errors.
\paragraph{Neural-symbolic Web World Model.}
\emph{AI Spire} separated creative generation and executable mechanics. The neural component (Gemini Flash) acts as a restricted designer: it receives the prompts and descriptions with the user's desired card power and mechanics and returns a JSON object with a name, card description text, and effect codes. The symbolic component in \texttt{App.tsx} acts as a rules engine: it translates these effect codes and applies them to the game status. For example, if some relic contains an effect like \texttt{start combat strength 1}, the trigger handler will detect the related events and increase the user's strength variable at the start of a combat. This kind of separation realized the safe creativity: the model could generate new things while the effects are restricted to the controlled vocabulary for code implementation.
\paragraph{Procedural content and The Wish.}
Standard rewards are generated by \texttt{generateRewardCards} under \texttt{geminiService.ts}, with the usage of LLM. Besides, the game also includes \emph{The Wish} mechanism, which allows the user to request a card with a free-form prompt, such as ``a fireball that could deal a large amount of burn but also freeze the enemy''. The \texttt{generateWishCard} service will translate the user prompt into effective mechanics. In this example, it will deal high burn (with an LLM-generated reasonable value) and also apply the freezing status to the enemy. A similar procedure works for the shop scene; when a player enters the shop, it will trigger the generation of a themed inventory, whose prices and rarity constraints are based on the current run state.
\paragraph{Robustness and fallbacks.}
The interface between generation and execution is guarded by schema validation. Under \texttt{geminiService.ts}, GenAI \texttt{responseSchema} is used shape the structure to meet the expectation (such as \texttt{CARD SCHEMA} and \texttt{RELIC SCHEMA}), and restrict the integer costs and valid card types (\texttt{ATTACK}, \texttt{SKILL}, and \texttt{POWER}). When missing valid API keys or the API call fails, the system will call the stored samples so that the gameplay will still be smooth.

\paragraph{Demonstrations.}
During a typical gameplay, the user may encounter different kinds of enemies, such as common/elite or boss; the game engine will take charge of bookkeeping, such as HP, block, status effects, and deck cycling. After winning a combat, the reward section will provide 3 real-time-generated cards, and also a box for the user to type their ``wish'' for a new card. The user may choose a reward from the 3 provided cards such as ``Echoing Blade'', which will ``deal 7 damage and also recover an \texttt{ATTACK} card from the discard pile'' to gather more resources, or he can customize his own reward by typing ``a card that will heal myself and restore energy'' so that he can gain HP when playing the generated card.

\subsection{A Sandbox Called AI Alchemy}
\begin{figure}
    \centering
    \includegraphics[width=0.8\linewidth]{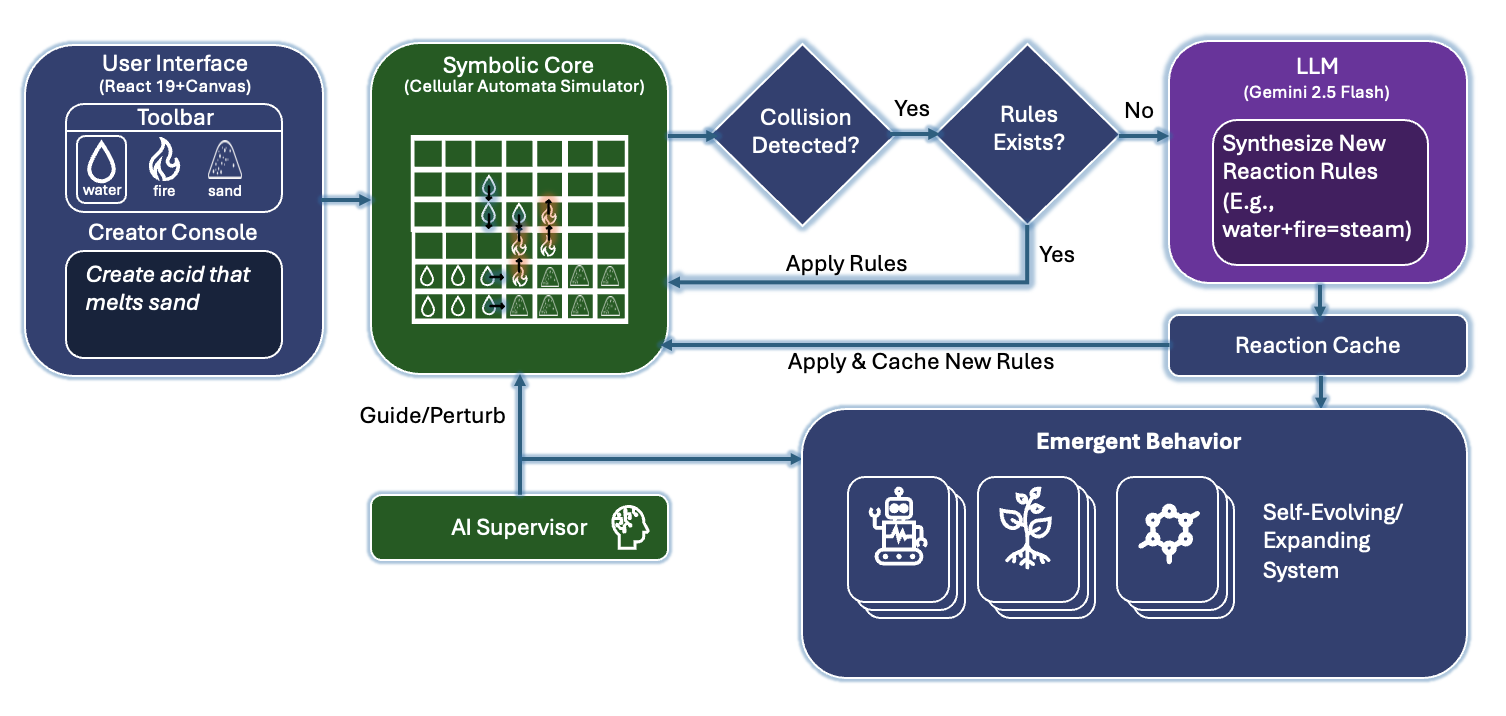}
    \caption{Neural-symbolic architecture of \emph{AI Alchemy}. A React+Canvas user interface (toolbar and natural-language Creator Console) injects user-defined materials into a symbolic cellular-automata ``falling-sand'' simulator. Upon particle collisions, the engine applies existing reaction rules when available; otherwise, it queries an LLM (Gemini Flash) to synthesize a schema-constrained reaction outcome, which is cached and immediately integrated into the update loop. An optional AI Supervisor monitors the canvas and guides/perturbs the system, enabling controlled emergent behavior in a self-expanding sandbox.}
    \label{fig:AI Alchemy}
\end{figure}

\begin{figure}
    \centering
    \includegraphics[width=0.8\linewidth]{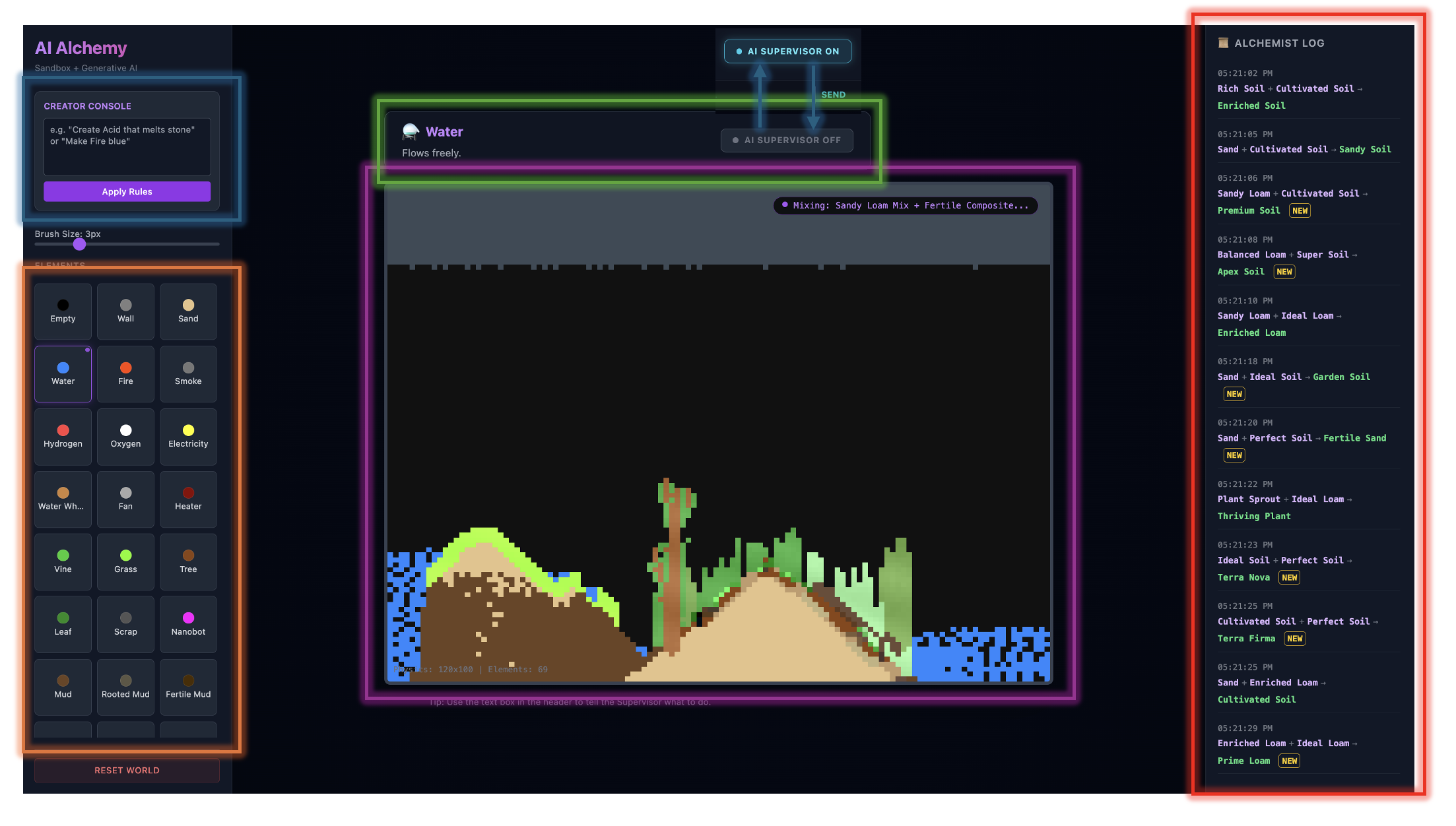}
    \caption{User interface of \emph{AI Alchemy}. The upper-left panel (blue box) shows the creator console, where users can use natural language to introduce new elements with desired rules. The lower-left panel (orange box) shows the element bar, displaying current available element. The top overlay (green box) indicates the AI supervisor layer, toggling whether LLMs are introduced into gameplay. The central canvas (purple box) represents the simulation canvas, where all the automata's behaviors take place. The right panel (red) displays the Alchemist log, recording LLM-generated elements and corresponding reactions.}
    \label{fig:AI Alchemy1}
\end{figure}

\emph{AI Alchemy} applies the WWM to the field of cellular automata, redefining the ``falling sand'' simulation genre. Traditional sandboxes rely on fixed reaction tables (e.g., \texttt{water+fire=steam}), limiting discoverability to developer presets. In \emph{AI Alchemy}, reactions and materials are open-ended: when a user introduces a new material or combination, the LLM proposes a valid reaction and result under physical constraints. This system utilizes real-time generation to autonomously expand the simulation ecosystem.

\paragraph{Environment and interaction.}
The interface utilizes React 19 and an HTML Canvas grid to simulate physics such as gravity, flow, and diffusion. Users select primary elements (Water, Fire, Sand) from a toolbar or use the \emph{Creator Console} to define new elements via natural language. The system translates these prompts into structured, physically-defined elements (with color, state, decay, and interaction rules). An optional \emph{AI Supervisor} acts as an autonomous agent, monitoring canvas state and perturbing the system to induce emergent behavior.

\paragraph{Neural-symbolic Web World Model.}
The symbolic core (\texttt{sandbox.tsx}) performs cellular automation, which relies on the physical categories, such as \texttt{POWDER}, \texttt{LIQUID}, and \texttt{GAS}, etc., to update the particles. The reaction is guided by \texttt{reactionCache/pendingResolution}: when two elements collide, the engine first checks if there are existing rules for the reaction. If not, the system will call the LLM and judge the reaction by the related element types, then produce a reaction outcome (such as Ash, Steam, etc.) as well as provide the constrained parameters for the simulator. The reaction result will be cached and immediately integrated into the automata mechanism. Safety and stability are enforced at the schema level: the generation can produce new concepts, but the simulator will restrict the rate limits (such as decay probability, energy use), to make the entire simulation under control.

\paragraph{Demonstrations.}
After replacing the preset reaction table with real-time rule generations, the system gets a bigger space for emergent behavior: during the gameplay, the system itself may explore rules like \texttt{Life+Fire=Ash}, \texttt{Ash+water=Nutrient mud}, while nutrient \texttt{mud+Life=more Life}. Even more complex behavior could also be simulated in the system, such as transport dynamics of nano-robots and machine-like elements like heaters and fans. The AI supervisor could avoid a single element dominating the world by monitoring the global statistics, then induce water into the system by rainfall, or destroy the dominated Life or elements by burning, or just directly send them to the void. Eventually, we'll get a sandbox system that is physically explainable while also being a self-expanding system with constrained generation.

\subsection{Cosmic Voyager}
\begin{figure}[ht!]
    \centering
    \includegraphics[width=0.8\linewidth]{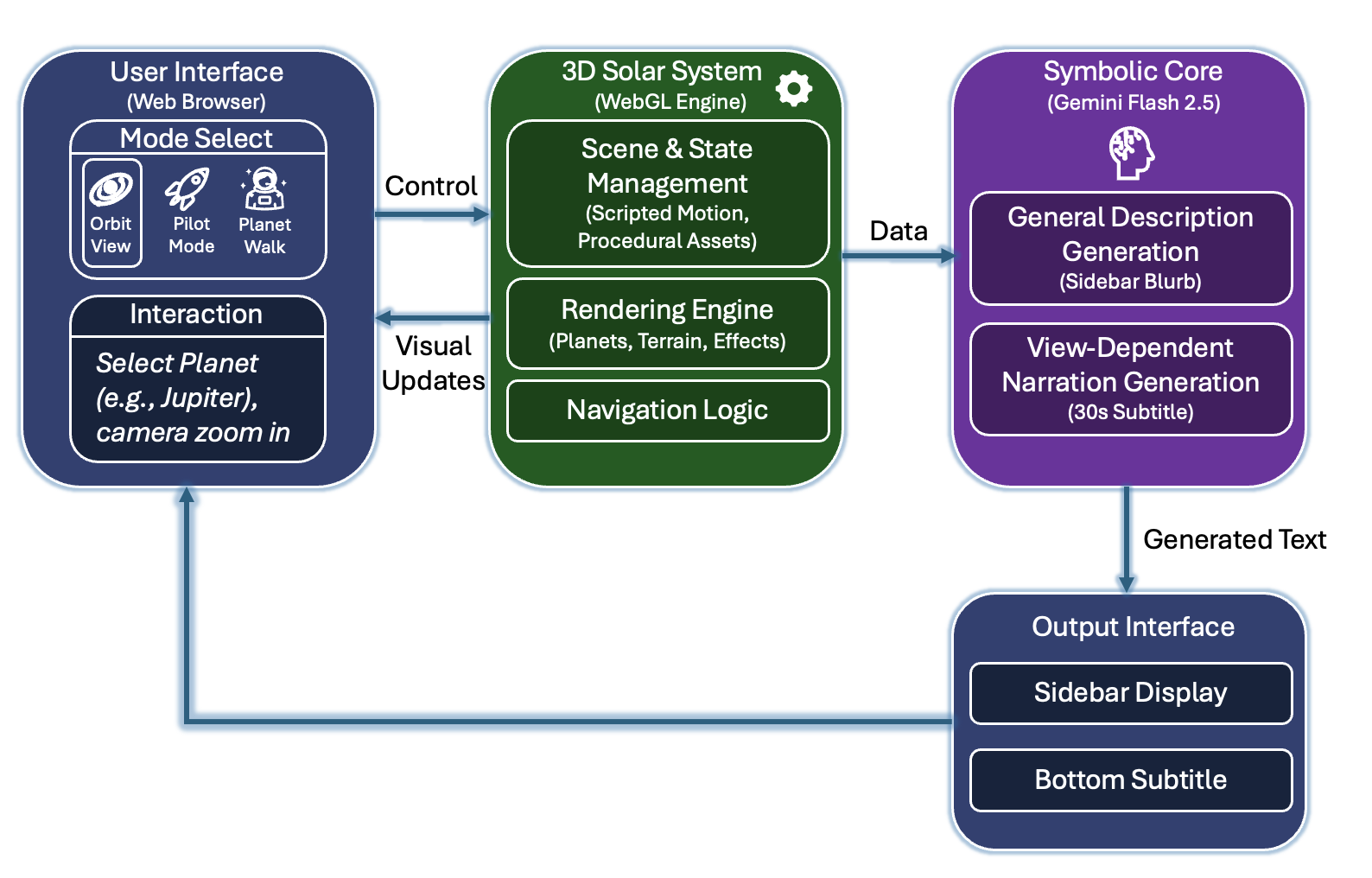}
    \caption{Neural-symbolic architecture of \emph{Cosmic Voyager}. A browser-based user interface (mode selection for orbit view, piloted flight, and surface walk, plus planet selection and camera control) drives a WebGL solar-system engine that manages scene/state, procedural assets, navigation logic, and rendering. The engine streams the currently selected celestial body and view context to a symbolic core (Gemini Flash), which generates (i) a short, general sidebar description and (ii) view-dependent ``Cosmic Guide'' narration that refreshes every 30\,s as a bottom subtitle with a typewriter reveal. When the API is unavailable, the system falls back to bundled descriptions to preserve a continuous educational experience.}
    \label{fig:Cosmic-Voyager1}
\end{figure}

\begin{figure}[ht!]
    \centering
    \includegraphics[width=0.8\linewidth]{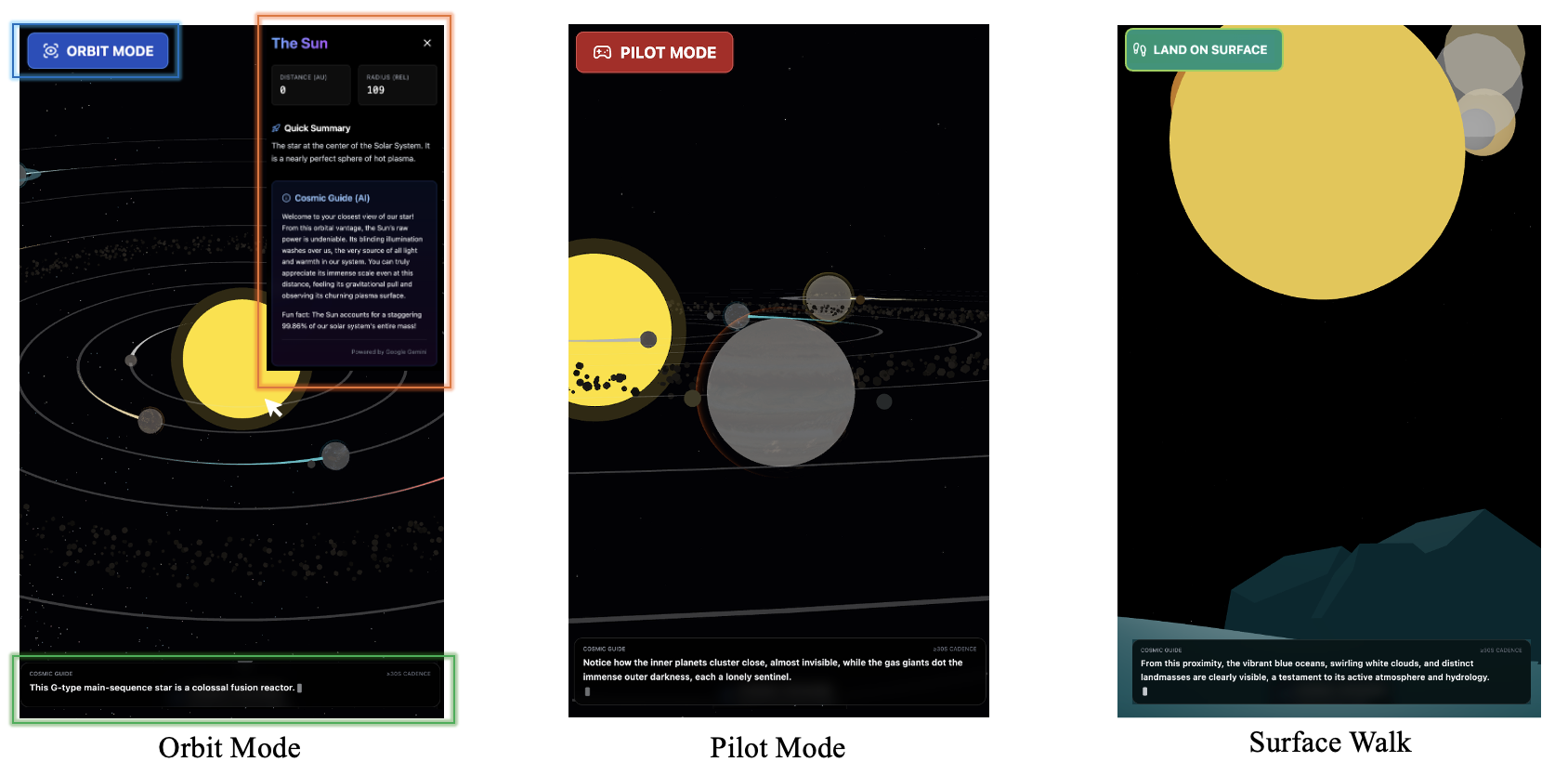}
    \caption{User interface of \emph{Cosmic Voyager} (left to right). \textbf{Orbit Mode} (circled with blue box) presents a high-level solar-system overview with selectable bodies; choosing an object (e.g., the Sun) opens a sidebar card (circled with orange box) with key stats and an AI-generated quick summary. \textbf{Pilot Mode} switches to a ship-like free-flight camera for navigable traversal through the system. \textbf{Surface Walk} enables first-person exploration on a generated planetary surface after landing. Across all modes, a persistent bottom ``Cosmic Guide'' (circled with green box) subtitle strip refreshes every 30~sec with view-dependent narration tied to the currently selected body and camera context.}
    \label{fig:Cosmic-Voyager2}
\end{figure}

Cosmic Voyager implements a 3D Web World Model for planetary exploration. Rather than relying on pre-rendered media, it provides a navigable solar system where spatial context dictates agent explanation. The experience functions as a lightweight spaceflight simulator: users switch between orbit viewing, piloted flight, and surface walking, while the AI guide generates educational narration matched to the current viewpoint.

\paragraph{Environment.}
Cosmic Voyager renders a stylized solar system using WebGL. The user can switch between multiple interaction modes: high-level orbital overview, a ship-like control scheme (piloted flight), and surface exploration on generated terrain. A bottom ‘Cosmic Guide’ subtitle strip auto-refreshes every 30 seconds, narrating the currently viewed body or context.

\paragraph{Neural-symbolic Web World Model.}Scene layout and motion are scripted for clarity rather than physical fidelity: preset orbital speeds, static distances, rim-glow atmospheres, optional rings, and procedural asteroid placement. Given the currently selected body’s name, a Gemini call returns a short, general description (with baked-in fallbacks if the API fails). In addition to a general description, a view-aware Cosmic Guide subtitle updates every 30s with Gemini narration tailored to the currently selected body and camera context.

\paragraph{Current capabilities and future extensions.}
Included content spans the Sun, major planets, their key moons, and playful claimed asteroids. Controls support orbiting, free flight, and pointer-locked walking atop a scaled sphere; the scales are intentionally compressed for usability. Live Gemini narration depends on supplying a \texttt{GEMINI\_API\_KEY}, without it, the app falls back to bundled descriptions. Potential extensions include richer surface variation, more realistic orbital mechanics, VR, or shared multiuser tours.

\paragraph{Demonstrations.}
A session opens in the orbital overview; selecting Jupiter triggers a smooth camera travel and surfaces an AI blurb in the sidebar. Clicking the asteroid belt snaps the focus to a nearby rock and shows the owner and mining information. Choosing to land on Mars switches to a first-person view on its spherical surface, with the Sun and other planets visible in the sky, while movement is simulated by rotating the world beneath the player.

\subsection{WWMPedia}

\begin{figure}[ht]
   \begin{subfigure}{0.33\textwidth}
       \includegraphics[width=\linewidth]{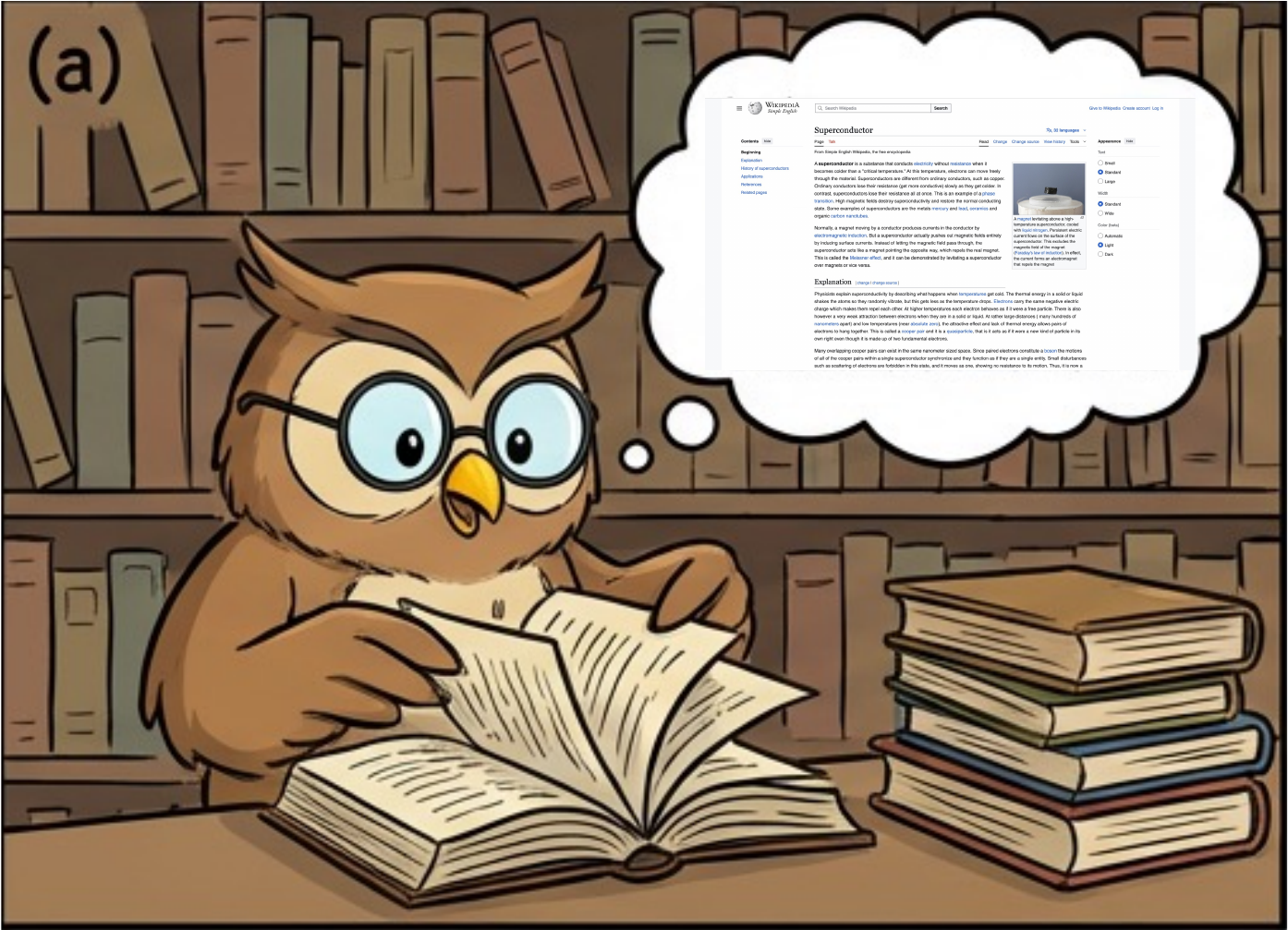}
       \caption{\textbf{Wikipedia} entry for \emph{Superconductor}.}
       \label{fig:wiki_superconductor}
   \end{subfigure}
\begin{subfigure}{0.33\textwidth}
       \includegraphics[width=\linewidth]{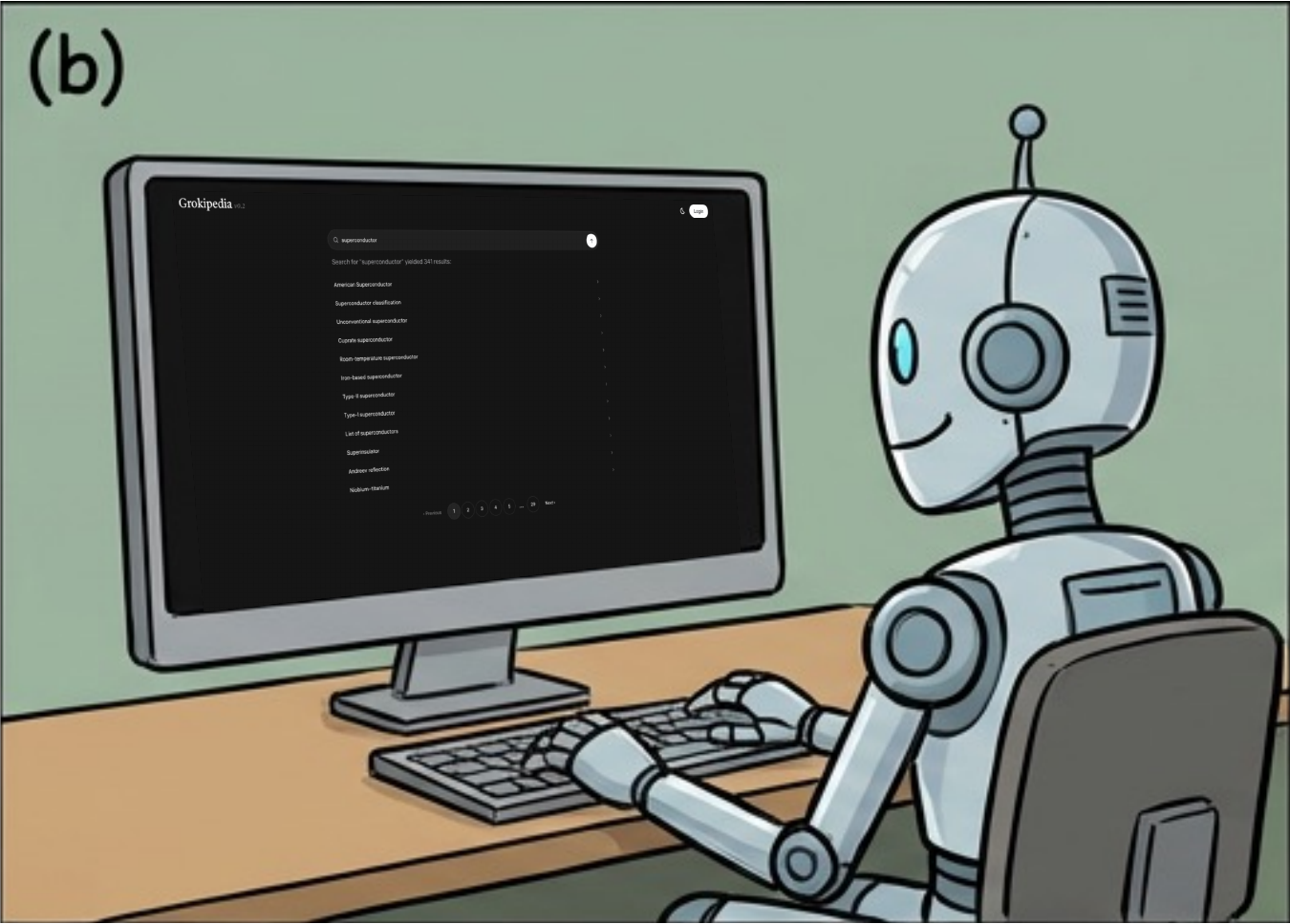}
       \caption{\textbf{Grokipedia} need to select from predefined articles.}
       \label{fig:grokpedia_superconductor}
   \end{subfigure}
   \begin{subfigure}{0.33\textwidth}
       \includegraphics[width=\linewidth]{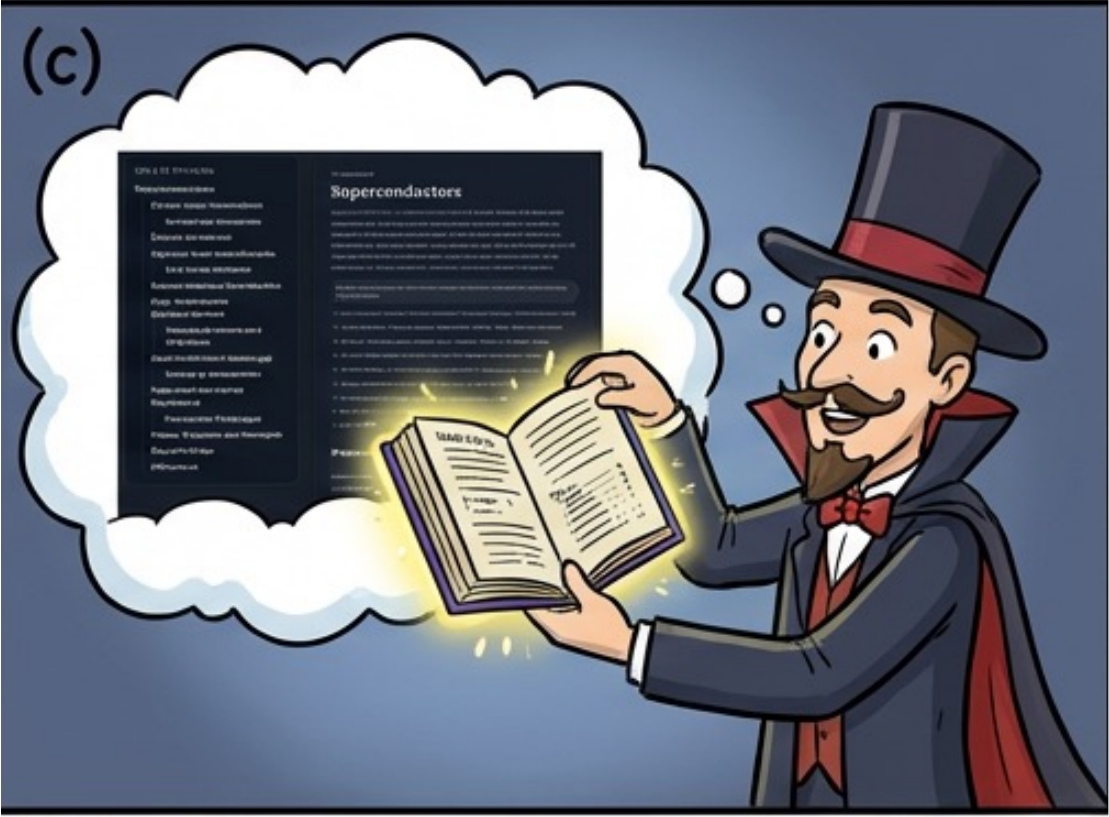}
       \caption{\textbf{WWMPedia} page generated on-demand for the same query.}
       \label{fig:WWM_superconductor}
   \end{subfigure}
   \caption{\textbf{Wikipedia vs.\ Grokipedia vs.\ WWMPedia} for the query ``Superconductor''. WWMPedia treats the open web as an unbounded knowledge world: the agent retrieves evidence via search and page opening (Physics, $S^{\phi}$), and an LLM composes a structured article view (Imagination, $S^{\psi}$) with a table of contents and sectioned prose, annotated with citations to the retrieved sources. In addition, the user could choose any section to be elaborated by clicking an "explain more" button. }
   \label{fig:WWMPedia_comparison}
\end{figure}

\noindent\textbf{WWMPedia} is our knowledge-centric Web World Model: instead of navigating a pre-indexed corpus, the user enters through a natural-language query, and the system synthesizes a Wikipedia-like page on the fly (similar to Grokpedia). Conceptually, the \emph{world} is the open web, and each generated page is a local ``state'' that makes this world legible and browsable.

\paragraph{Environment.}
WWMPedia's environment is the live web, exposed through a small set of browser primitives: (i) search for a query, (ii) open candidate pages, and (iii) extract text spans as evidence. This environment is effectively unbounded in topic space and partially observable in practice: the agent only sees what it chooses to retrieve, and different queries surface different neighborhoods of the web.

\paragraph{WWM instantiation.}
WWMPedia is a lightweight instantiation of the WWM split $S_t=(S^{\phi}_t,S^{\psi}_t)$. The Physics layer $S^{\phi}$ is implemented as ordinary web scaffolding: query routing, retrieval, sanitization, and a deterministic HTML renderer that enforces a fixed page layout (title, table of contents, sections, and references). The Imagination layer $S^{\psi}$ is produced by the LLM: given the retrieved evidence bundle, it selects an outline, writes a sectioned exposition, and emits a reference list that links generated statements back to sources. The UI surfaces this provenance explicitly (e.g., ``LLM generated'' and a generation timestamp in Figure~\ref{fig:WWM_superconductor}), which makes the page feel more like a stateful artifact than a simple chat response.

\paragraph{Why this is a WWM.}
From the WWM perspective, the key design choice is that \emph{retrieval and rendering are code-defined}. The model operates inside a structured interface defined by web code, and its output is shaped into a stable, inspectable page. Compared to Grokipedia, where the user has to choose a predefined entry from a dropdown menu, our WWMPedia removes this restriction and generates everything on the fly. Each section is not fixed as well, as they could be further elaborated upon the user's request.

\subsection{Bookshelf}

\begin{figure}[ht]
   \centering
   \begin{subfigure}{0.95\textwidth}
       \includegraphics[width=\linewidth]{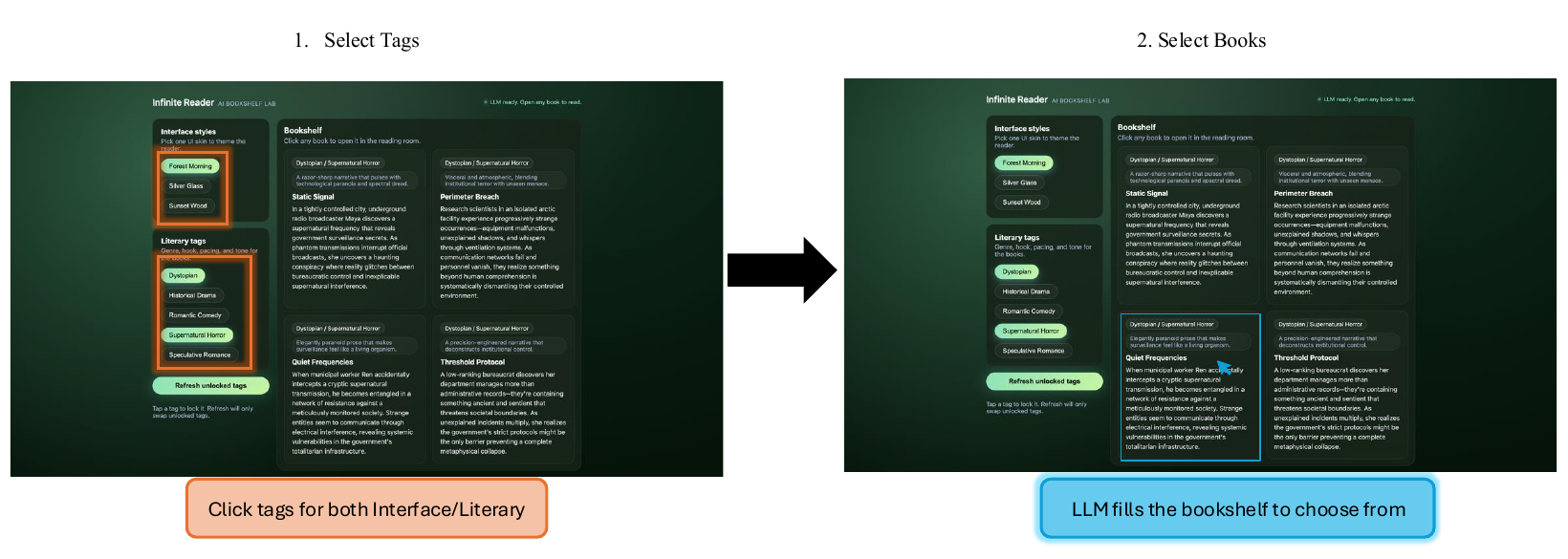}
       \caption{\textbf{Library view.} Users pick a UI skin (interface style) and literary tags, then select from LLM-proposed book cards.}
       \label{fig:bookshelf_entry}
   \end{subfigure}
   \begin{subfigure}{0.95\textwidth}
       \includegraphics[width=\linewidth]{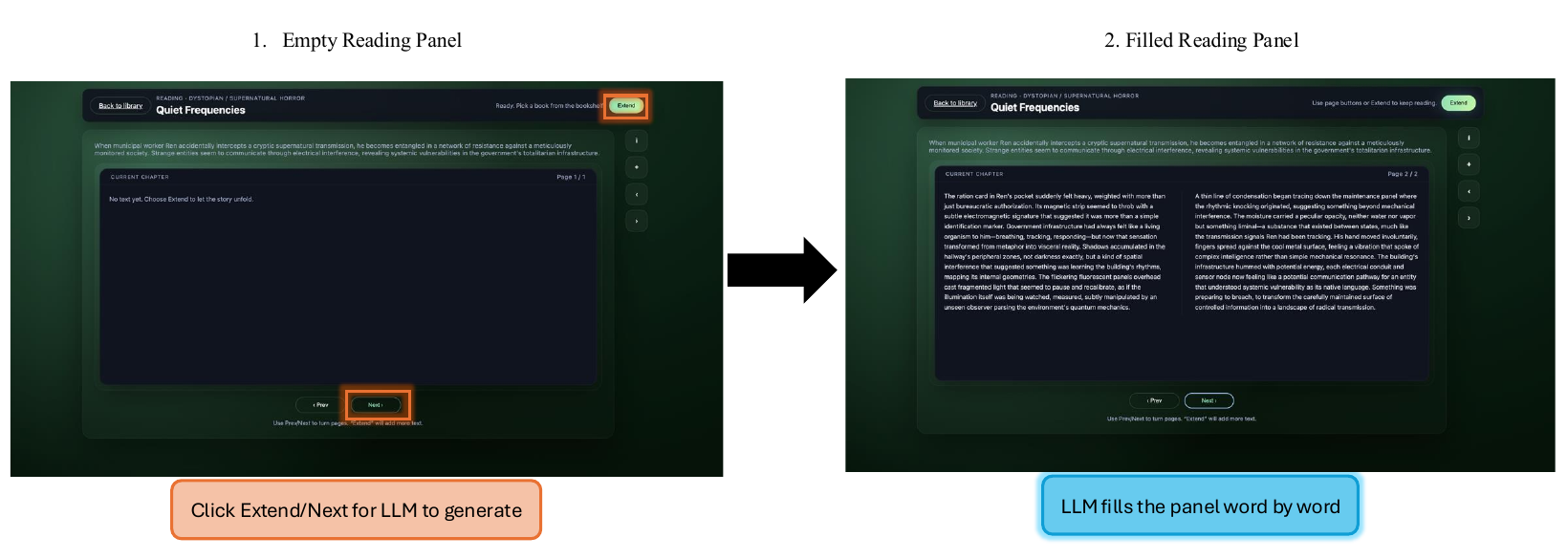}
       \caption{\textbf{Reading view.} Page-turn actions (Prev/Next) and \texttt{Extend} stream new text while preserving the active style constraints.}
       \label{fig:bookshelf_reader}
   \end{subfigure}
   \caption{\emph{Bookshelf} (\emph{Infinite Reader}) interface. The user controls generation through a compact, typed control surface: interface-style tags deterministically theme the UI, while literary tags constrain genre, tone, and pacing of the narrative.}
   \label{fig:bookshelf_ui}
\end{figure}

Bookshelf explores WWMs in a different regime: long-form generative fiction. The user interaction loop is intentionally minimal (select a book, then turn pages), but the system must still maintain continuity, constrain style, and present content in a stable, readable interface.

\paragraph{Interface and interaction.}
The entry screen (Figure~\ref{fig:bookshelf_entry}) exposes two orthogonal tag sets that function as a compact control language. \emph{Interface styles} select a UI skin (typography, spacing, and palette), and are implemented as deterministic CSS/theming choices in the client. \emph{Literary tags} specify narrative constraints (genre, tone, pacing), and condition both the proposed book cards (title/tagline/blurb) and subsequent page generations. The ``Refresh unlocked tags'' mechanism keeps the shelf feeling dynamic: some tags can be held fixed as anchors, while others are rotated to explore adjacent styles without requiring users to author prompts.

\paragraph{Latent state and generation loop.}
Under the hood, Bookshelf is a straightforward WWM where the Physics layer is narrative \emph{mechanics} rather than spatial dynamics. Source code defines the session state and page-turn semantics: page length limits, streaming boundaries, UI component composition, and which fields are carried forward across turns. On each page-turn, the system prompts the LLM with (i) the active tag constraints, (ii) the compact story state, and (iii) a short window of recent text. The LLM returns a continuation that is streamed into the reading panel (Figure~\ref{fig:bookshelf_reader}). 
\paragraph{Controllability and persistence.}
Bookshelf highlights a practical engineering constraint that appears across WWMs: long-horizon generation is mostly a \emph{state management} problem. We found it useful to keep the carried state typed and small. The LLM is responsible for local prose and scene-level detail, while code preserves invariants such as stylistic constraints, pagination, and what the system believes are the current open plot threads. This split gives a stable interaction contract even as the story world expands indefinitely.

\section{Related Work}

The development of persistent, intelligent environments involves integration of the advancements in world modeling, language agent architectures, and neuro-symbolic reasoning. We categorize the current literature into four main research streams: foundational world models, persistent agent environments, dynamic game generation, and agent reasoning frameworks.

\paragraph{World Models and Web Architectures.}
The modern concept of World Models was introduced in 2018 by \citet{ha2018worldmodels}, who showed that the agent could learn the policy evolutions entirely inside the dream environment, which was generated by the recurrent neural network. With the advent of large language models (LLMs), recent research has shifted towards utilizing the inherent knowledge of pre-trained transformers as world simulators. \citet{gu2024secretly} discussed if LLMs could act as internet world models, and propose \textit{WebDreamer}, which uses an LLM to simulate and score candidate action outcomes before execution, thereby reducing the amount of risky live exploration required for planning. Similarly, \citet{hao2023reasoning} proposed \textit{Reasoning via Planning (RAP)}, which could act as the LLM as world model and reasoning agent at the same time, such that it could run Monte Carlo Tree Search in the latent space of language. Bridging these generative approaches with practical engineering draws on foundational ideas from the early days of neural networks: \citet{schmidhuber1990line} describes an online method where a recurrent model of environment dynamics (including future reinforcement) provides a differentiable pathway for credit assignment, and it discusses how such a model can also support planning over future action sequences. This line of work was later expanded with hierarchical temporal abstraction via history compression, which learn multi-timescale predictive representations for long-horizon sequence modeling \citet{schmidhuber1992learning}. More recently, \citet{lecun2022path} has advocated for predictive world models as a cornerstone of autonomous intelligent systems, arguing that agents should continually learn hierarchical latent models of the world to facilitate model-based reasoning and planning. Concretely, LeCun’s predictive-world-model agenda has been instantiated through Joint-Embedding Predictive Architectures that predict missing information in an abstract representation space, e.g., I-JEPA for images \citet{assran2023self} and V-JEPA for videos \citet{bardes2024revisitingfeaturepredictionlearning}. In parallel, diffusion-based generators provide an alternative route to learned simulators; diffusion transformers (DiT) offer a scalable backbone for diffusion models used in high-fidelity visual generation \citet{peebles2023scalable}.

\paragraph{Persistent Agent Environments and Social Simulacra.}
Creating environments that support long-term agent interaction requires robust memory and social simulation capabilities. \citet{park2023generative} pioneered the \textit{Generative Agents} architecture, which utilizes a memory stream and reflection mechanism to simulate believable human behavior and social emergence in a sandbox environment. This work builds upon their earlier exploration in \textit{Social Simulacra} \citep{park2022social}, which used LLMs to prototype social computing systems by populating them with diverse simulated user personas. In the domain of embodied agents, \citet{wang2023voyager} introduced \textit{Voyager}, an LLM-powered agent in Minecraft that continuously learns by curating a library of executable code skills, enabling open-ended exploration. Concurrently, \citet{zhu2023ghost} proposed \textit{Ghost in the Minecraft (GITM)}, which employs hierarchical planning and text-based knowledge retrieval to handle long-horizon tasks in complex open worlds.

\paragraph{Dynamic Games and Neuro-Symbolic AI.}
The intersection of generative AI and interactive fiction presents unique challenges in state consistency and novelty adaptation. \citet{li2024unbounded} pushed the boundaries of this field with \textit{Unbounded}, a generative infinite game that simulates character life with open-ended interaction using specialized distillation techniques. To address the consistency issues in such open worlds, neuro-symbolic approaches have gained traction. \citet{balloch2023neurosymbolic} proposed neuro-symbolic world models that leverage symbolic graphs to adapt rapidly to open-world novelty. Earlier work by \citet{ammanabrolu2019playing} demonstrated the efficacy of graph-based deep reinforcement learning in text-adventure games, utilizing knowledge graphs to track state changes. Furthermore, \citet{huang2022language} explored the potential of LLMs as zero-shot planners, extracting actionable knowledge directly from language models to guide embodied agents in interactive environments.

\paragraph{Agent Reasoning, Learning, and Benchmarks.}
Robust agent performance in these environments relies on advanced reasoning and evaluation frameworks. \citet{yao2023react} introduced \textit{ReAct}, a paradigm synergizing reasoning and acting to improve task solving. To enhance long-term adaptability, \citet{shinn2023reflexion} proposed \textit{Reflexion}, allowing agents to learn from verbal reinforcement feedback, while \citet{majumder2023clin} developed \textit{CLIN}, a continually learning agent that abstracts causal models from interaction history. Evaluating these capabilities requires comprehensive benchmarks: \citet{zhou2023sotopia} focused on social intelligence through interactive scenarios, \citet{wu2023smartplay} provided a suite of games to test general intelligent capabilities, and \citet{lin2023agentsims} offered an open-source sandbox for evaluating planning and tool-use in simulated societies. Recent visual-spatial benchmarks further probe whether multimodal models build internal “local world models,” and show gains when models explicitly construct cognitive maps for downstream reasoning \citet{yang2025thinking,yin2025spatial}.

\section{Conclusion}

In this work, we introduce the \textbf{Web World Model} (WWM), an architectural paradigm that bridges the dichotomy between fixed-context web frameworks and unconstrained generative environments. By explicitly decoupling deterministic code, which governs state transitions and physical invariants, from the probabilistic creativity of large language models, we demonstrate a path toward scalable, hallucination-free worlds that do not rely on static databases. Our suite of applications validates that standard web protocols, when combined with typed latent interfaces and procedural hashing, serve as a robust substrate for persistent, open-ended exploration. Ultimately, WWMs establish a practical middle ground, enabling developers to build environments where language agents can act with structural certainty while retaining the capacity for unlimited imagination.

\vspace{5ex}
\bibliographystyle{plainnat}
\bibliography{reference}

\clearpage
\appendix

\renewcommand{\appendixpagename}{\centering \huge Appendix}
\appendixpage
\counterwithin{theorem}{section}

\startcontents[section]
\printcontents[section]{l}{1}{\setcounter{tocdepth}{2}}
\vspace{15ex}

\section{Additional Examples}

This appendix provides a comprehensive visual gallery of the implemented Web World Models. We showcase the UI states, generative diversity, and interaction flows for the \emph{Infinite Travel Atlas}, \emph{Galaxy Travel Atlas}, and other demonstrations discussed in the main text.

\begin{figure}[ht!]
  \centering
  \includegraphics[width=0.6\linewidth]{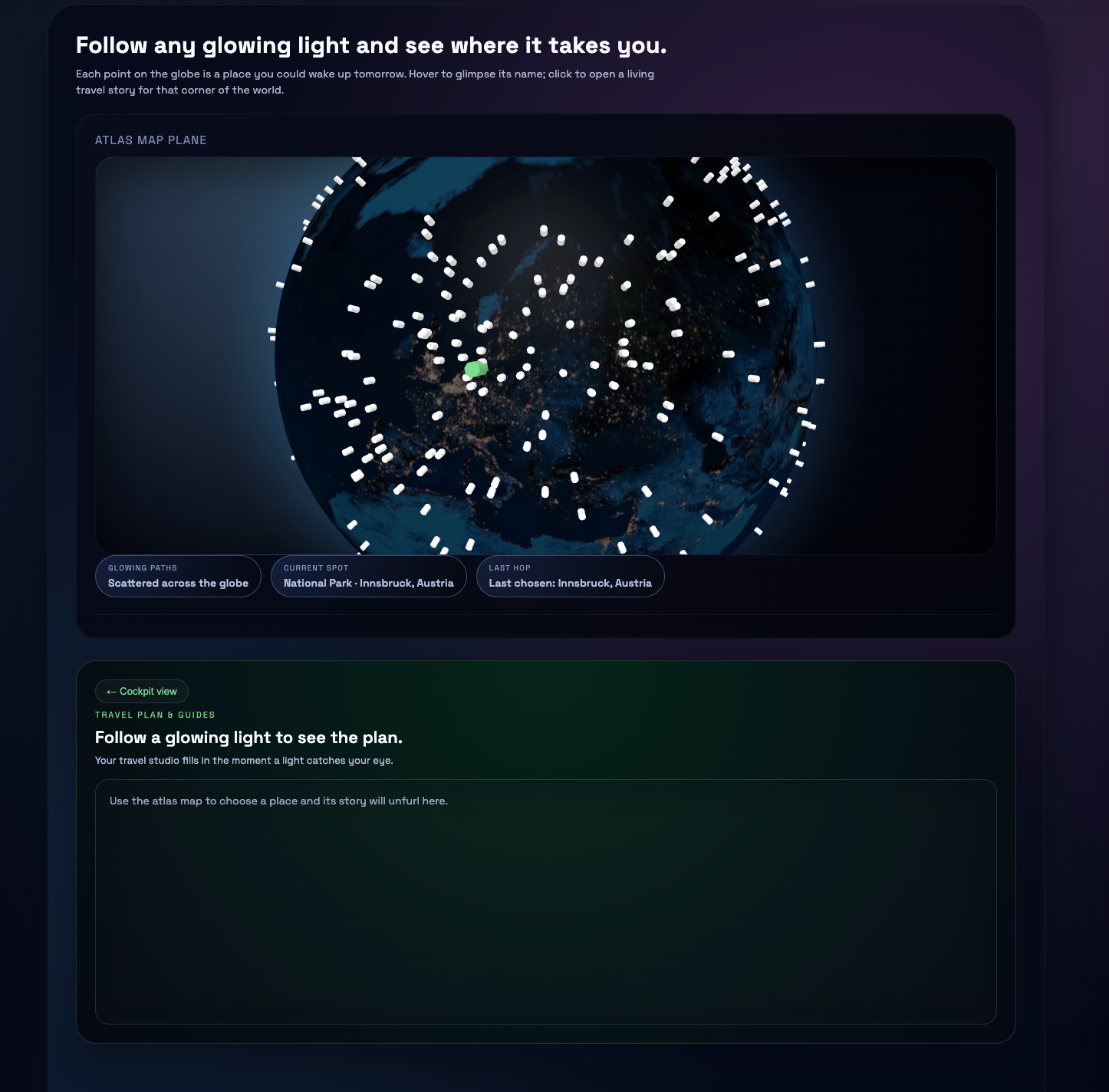}
  \caption{\textbf{Infinite Travel Atlas (a)} Initial globe view with an empty cockpit inviting the user to pick any glowing beacon.}
  \label{fig:atlas-blank}
\end{figure}

\begin{figure}[ht!]
  \centering
  \includegraphics[width=0.6\linewidth]{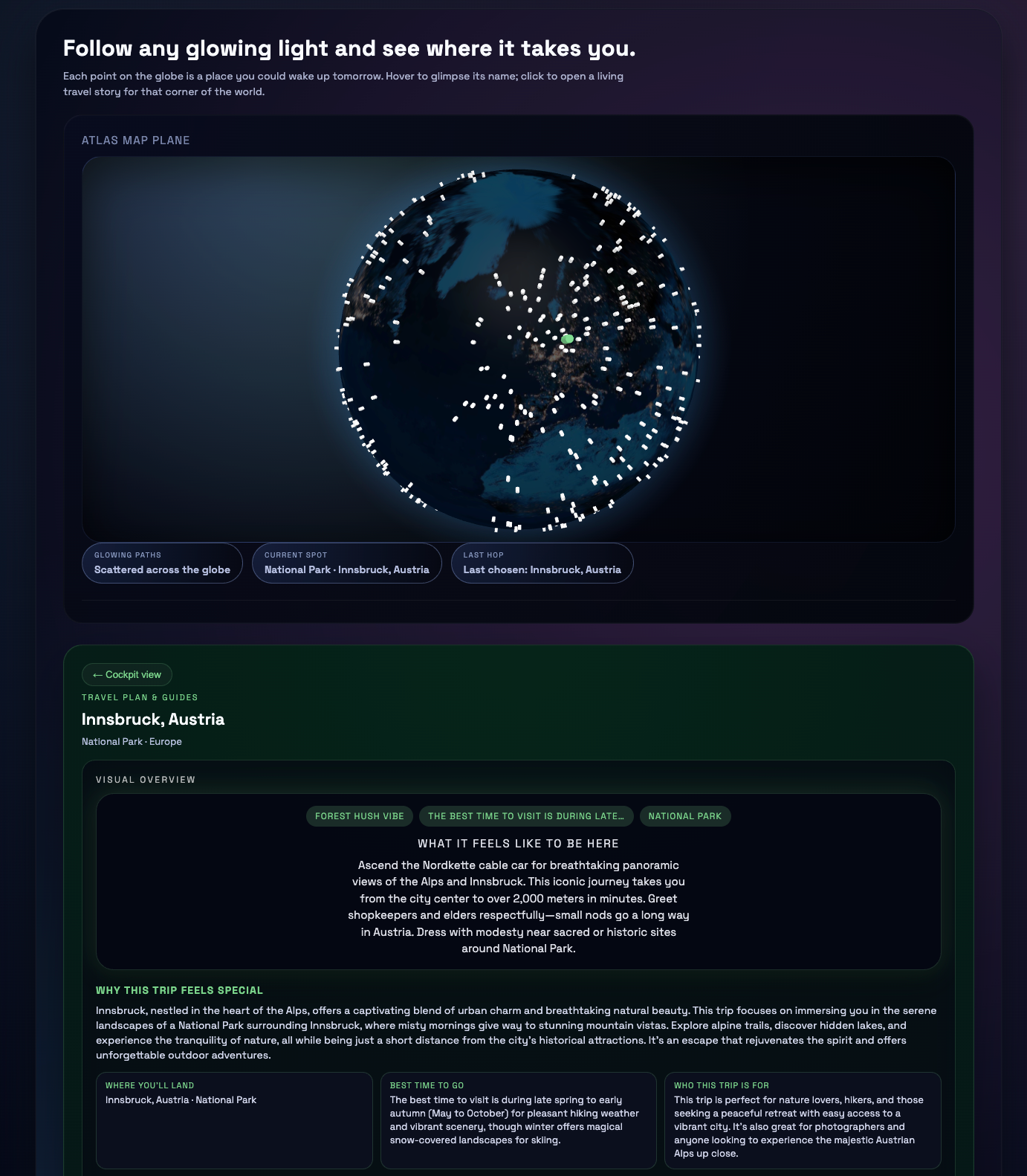}
  \caption{\textbf{Infinite Travel Atlas (b)} After selecting Innsbruck, the beacon lights up and the cockpit begins to fill with a themed guide.}
  \label{fig:atlas-preview}
\end{figure}

\begin{figure}[ht!]
  \centering
  \includegraphics[width=0.6\linewidth]{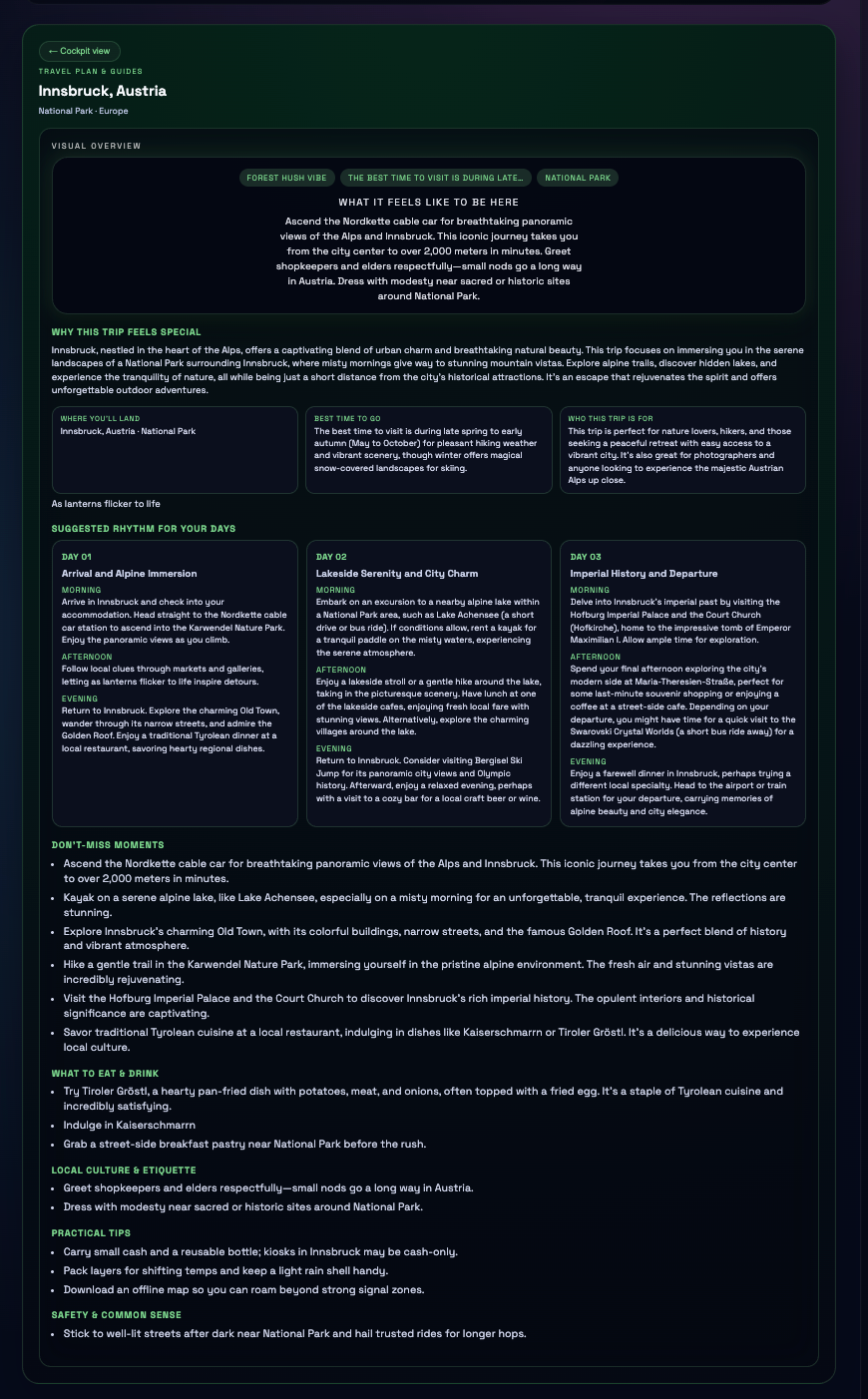}
  \caption{\textbf{Infinite Travel Atlas (c)} Full generated guide for Innsbruck rendered as a scrollable cockpit view.}
  \label{fig:atlas-innsbruck}
\end{figure}

\begin{figure}[ht!]
  \centering
  \includegraphics[width=0.6\linewidth]{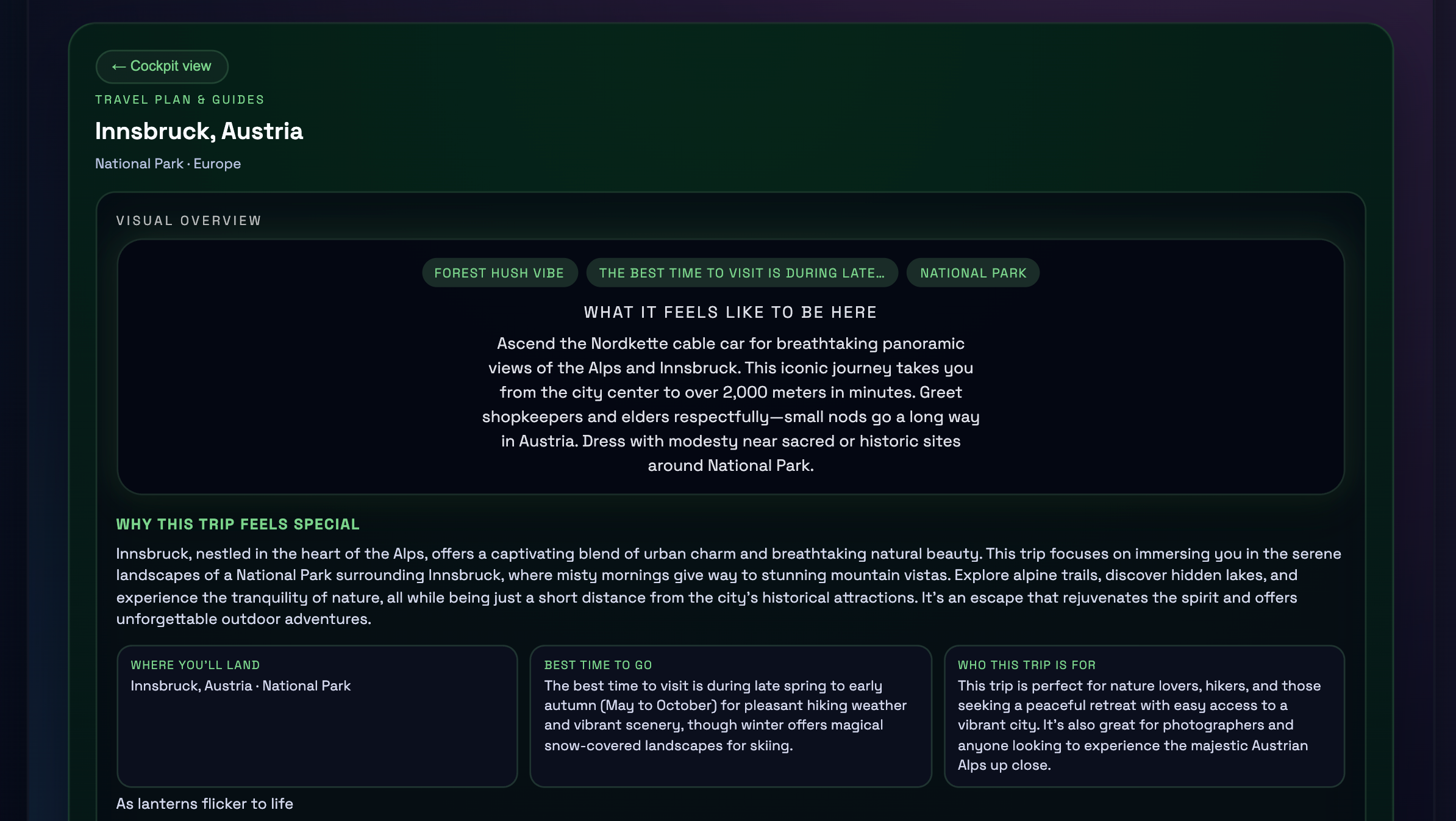}
  \caption{\textbf{Infinite Travel Atlas (d)} Top section of the Innsbruck guide: visual overview and why this trip feels special.}
  \label{fig:atlas-innsbruck-top}
\end{figure}

\begin{figure}[ht!]
  \centering
  \includegraphics[width=0.6\linewidth]{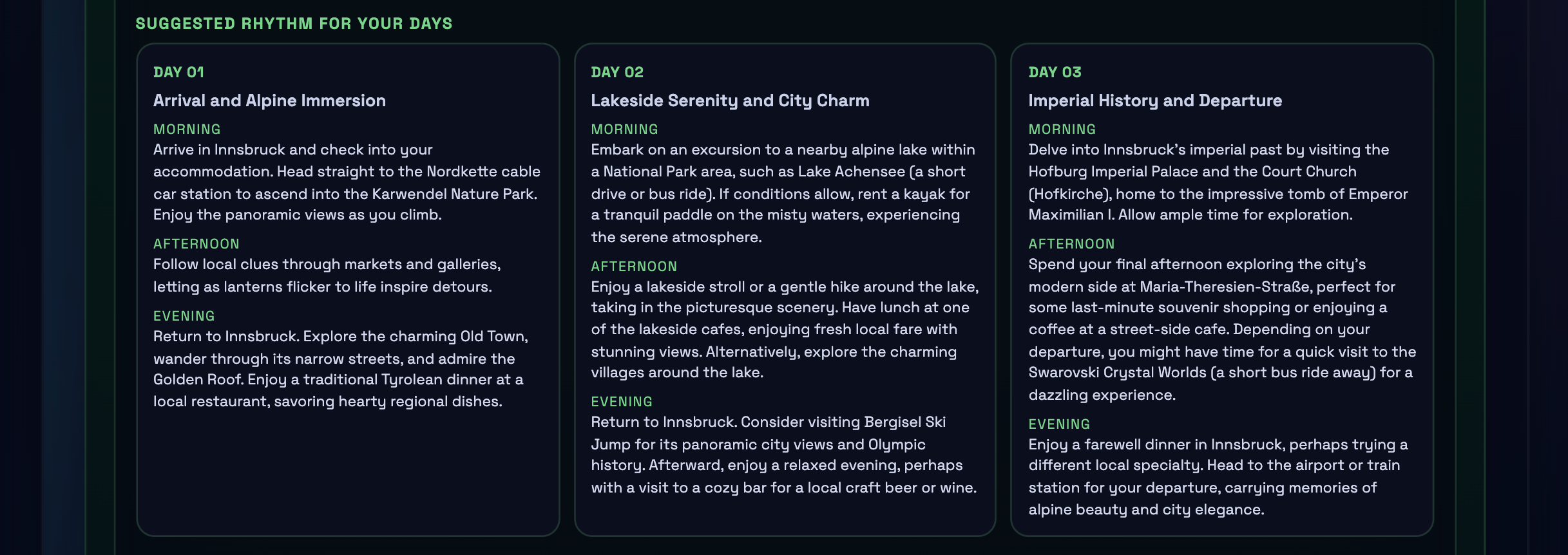}
  \caption{\textbf{Infinite Travel Atlas (e)} Mid-section: a three-day suggested rhythm with morning/afternoon/evening cards.}
  \label{fig:atlas-innsbruck-rhythm}
\end{figure}

\begin{figure}[ht!]
  \centering
  \includegraphics[width=0.6\linewidth]{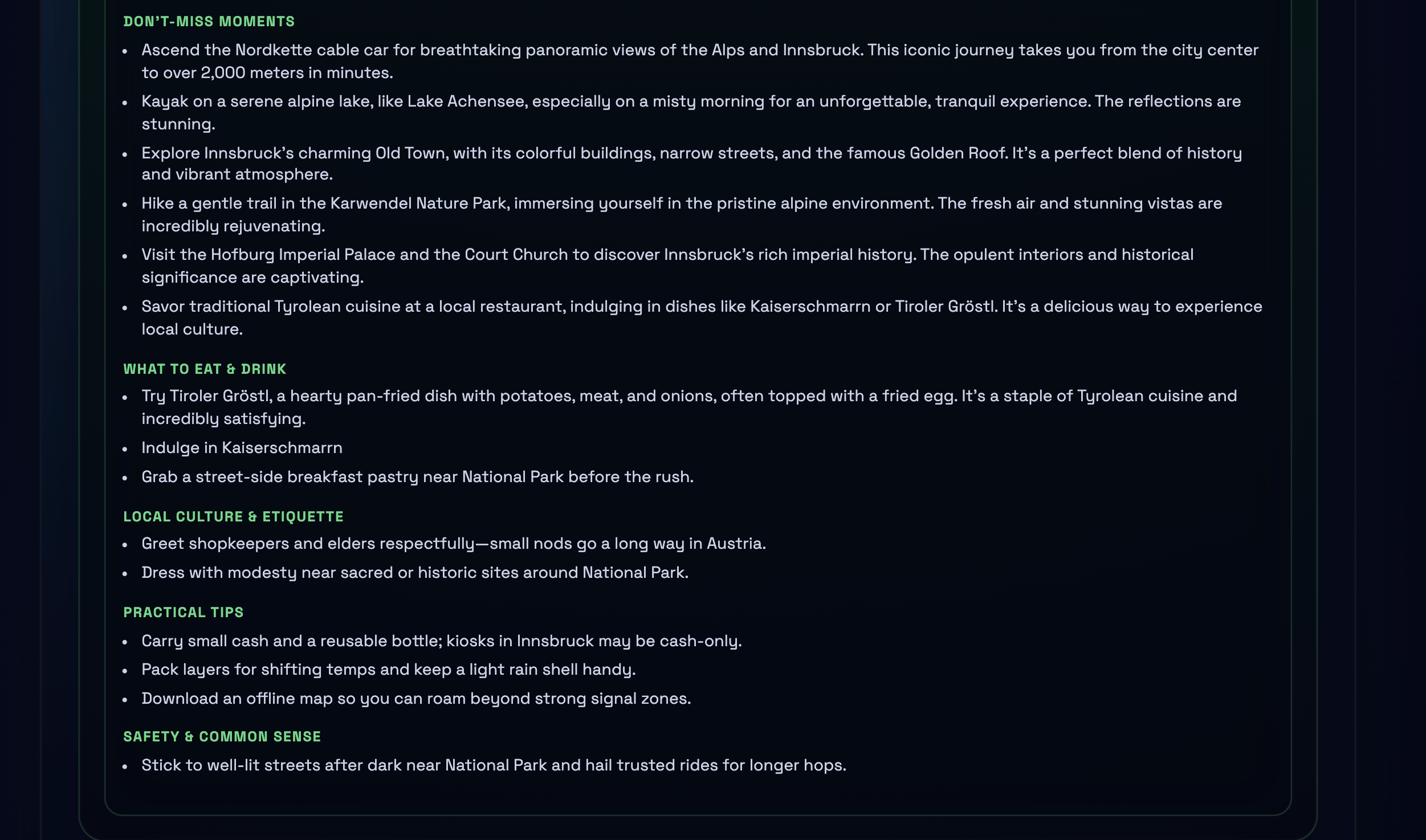}
  \caption{\textbf{Infinite Travel Atlas (f)} Bottom sections: don't-miss moments, food, culture, practical tips, and safety guidance.}
  \label{fig:atlas-innsbruck-bottom}
\end{figure}

\begin{figure}[ht!]
  \centering
  \includegraphics[width=0.6\linewidth]{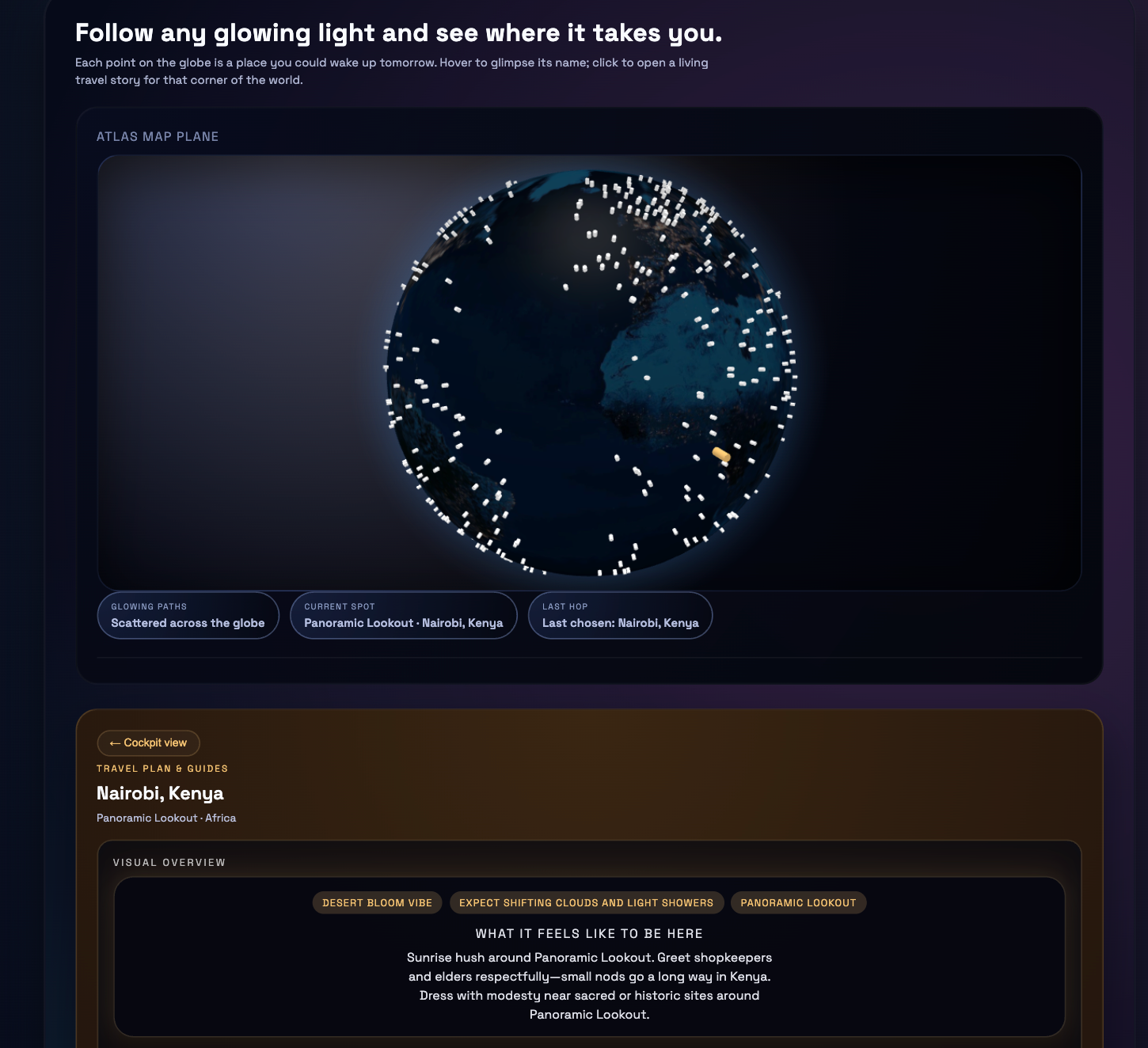}
  \caption{\textbf{Infinite Travel Atlas (g)} Nairobi, Kenya: warm desert-bloom theme and amber cockpit for a panoramic lookout.}
  \label{fig:atlas-nairobi}
\end{figure}

\begin{figure}[ht!]
  \centering
  \includegraphics[width=0.6\linewidth]{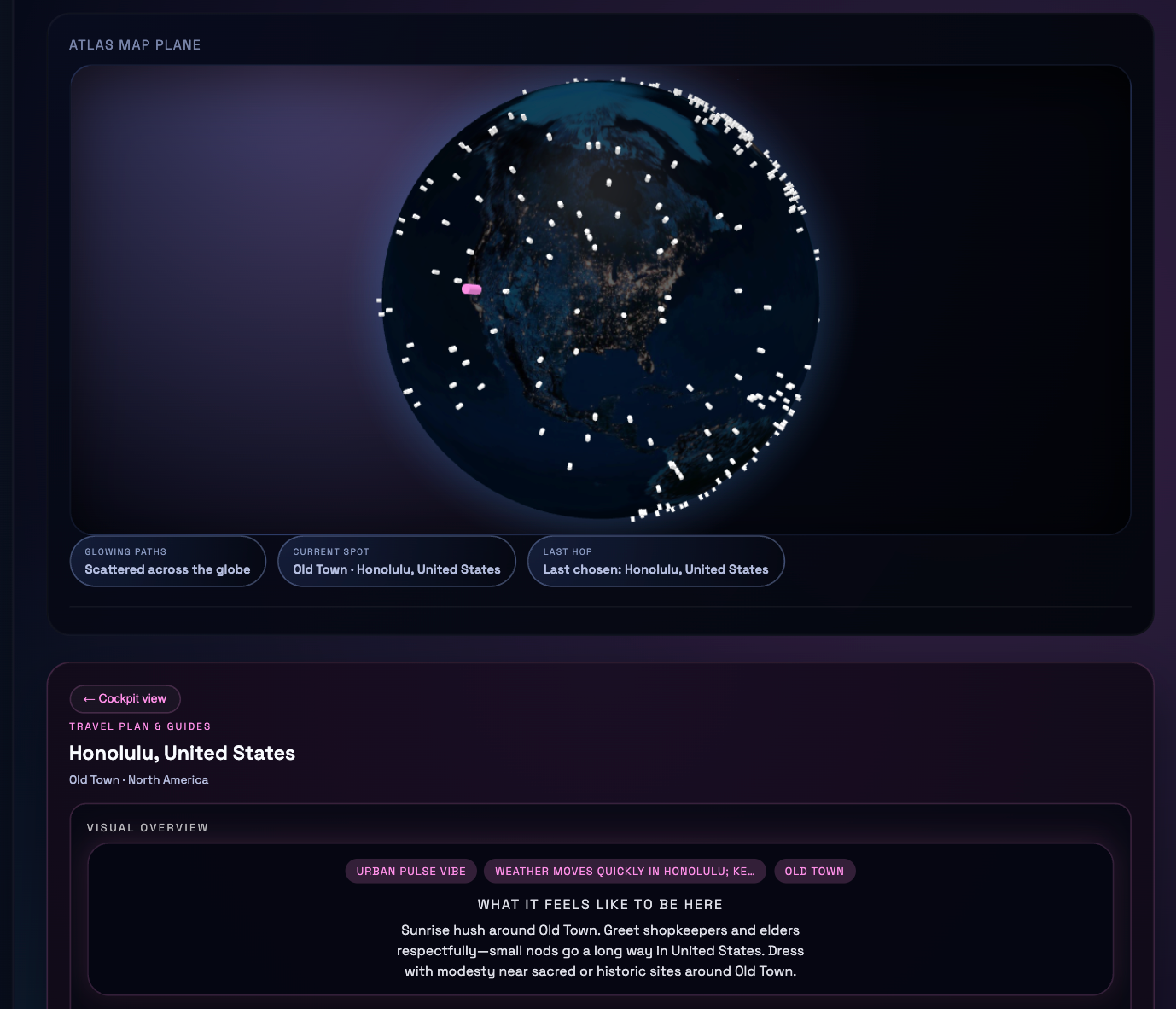}
  \caption{\textbf{Infinite Travel Atlas (h)} Honolulu, United States: urban-pulse theme with a violet palette for an old-town district.}
  \label{fig:atlas-honolulu}
\end{figure}

\begin{figure}[ht!]
  \centering
  \includegraphics[width=0.6\linewidth]{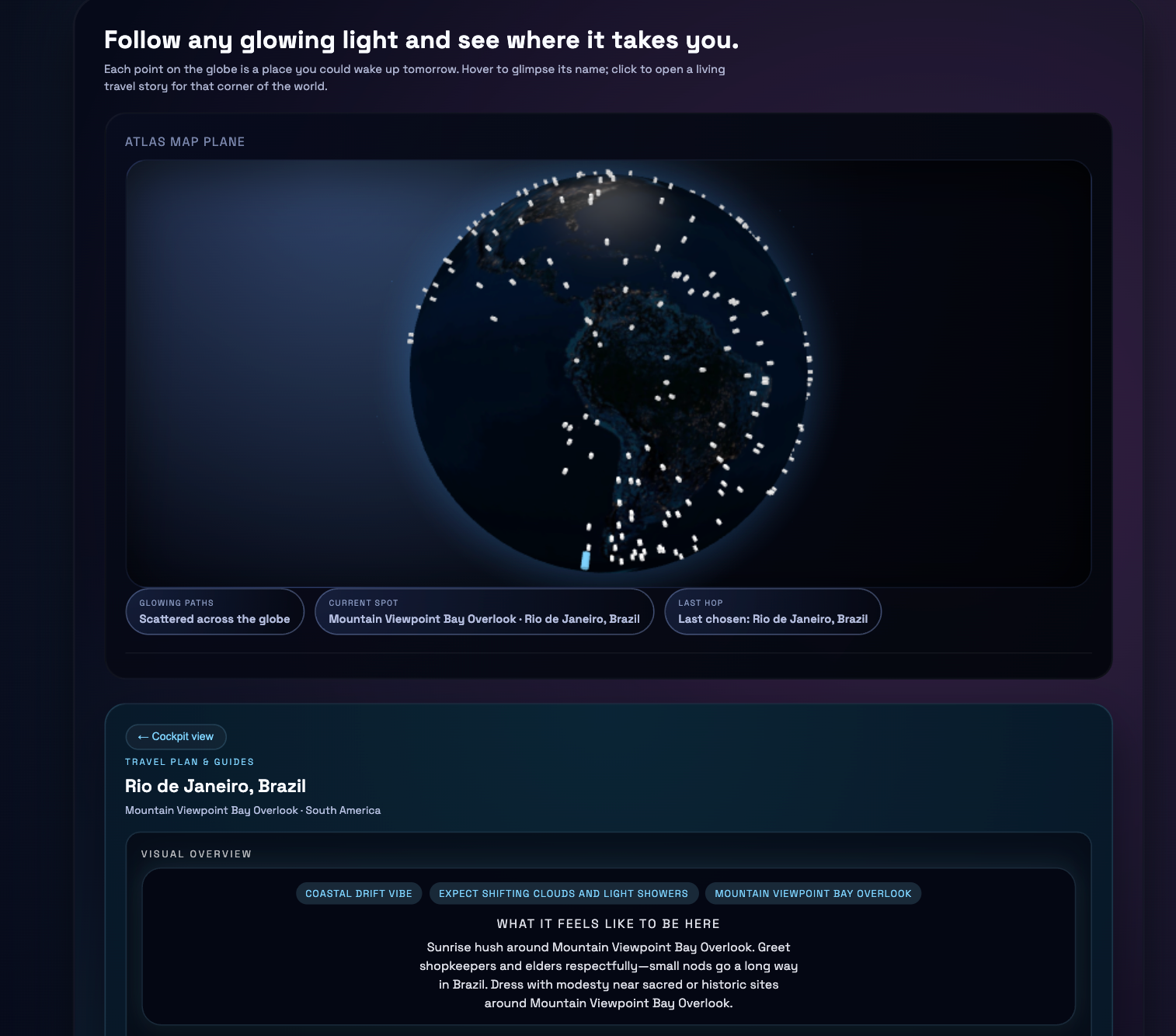}
  \caption{\textbf{Infinite Travel Atlas (i)} Rio de Janeiro, Brazil: coastal-drift theme and blue cockpit for a bay-overlook vantage point.}
  \label{fig:atlas-rio}
\end{figure}

\begin{figure}[ht!]
  \centering
  \includegraphics[width=0.6\linewidth]{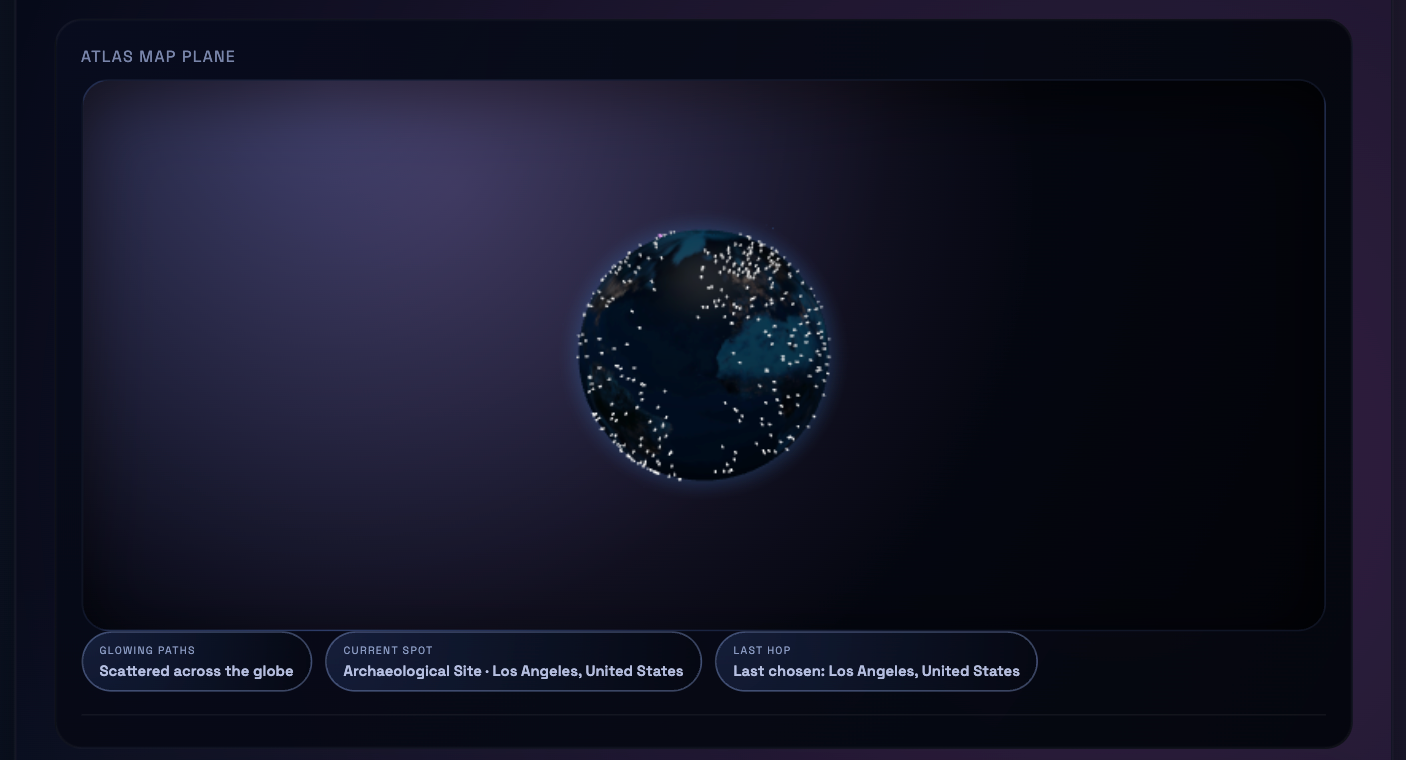}
  \caption{\textbf{Infinite Travel Atlas (j)} Los Angeles, United States: compact globe framing and atlas labels for an archaeological-site hop.}
  \label{fig:atlas-la}
\end{figure}

\begin{figure}[ht!]
  \centering
  \includegraphics[width=0.6\linewidth]{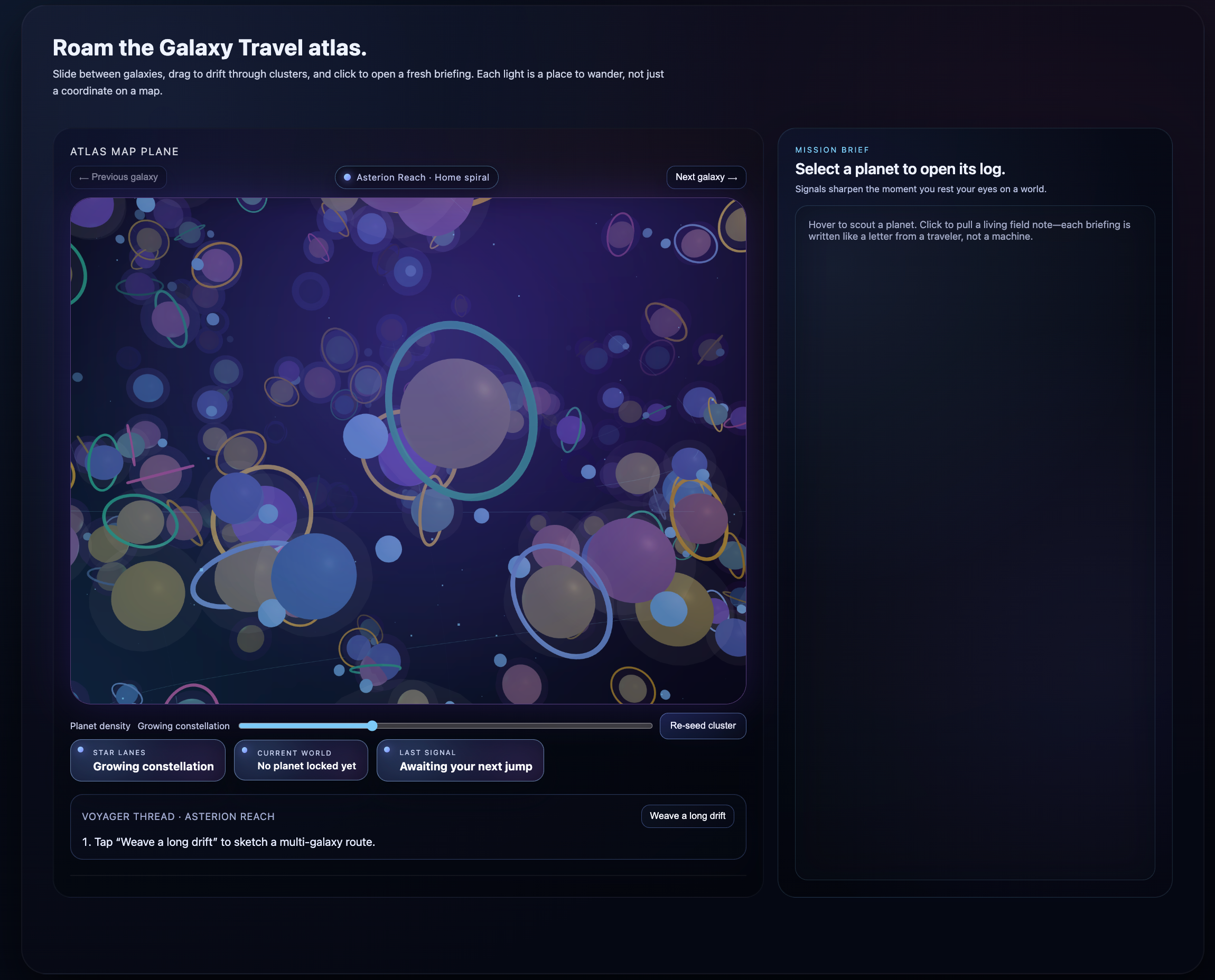}
  \caption{Galaxy Travel Atlas landing state. The left map plane is already populated by a procedurally generated cluster, while the mission brief panel stays empty and prompts the user to select a planet to open its log.}
  \label{fig:sci-fi-demo1-landing}
\end{figure}

\begin{figure}[ht!]
  \centering
  \includegraphics[width=0.6\linewidth]{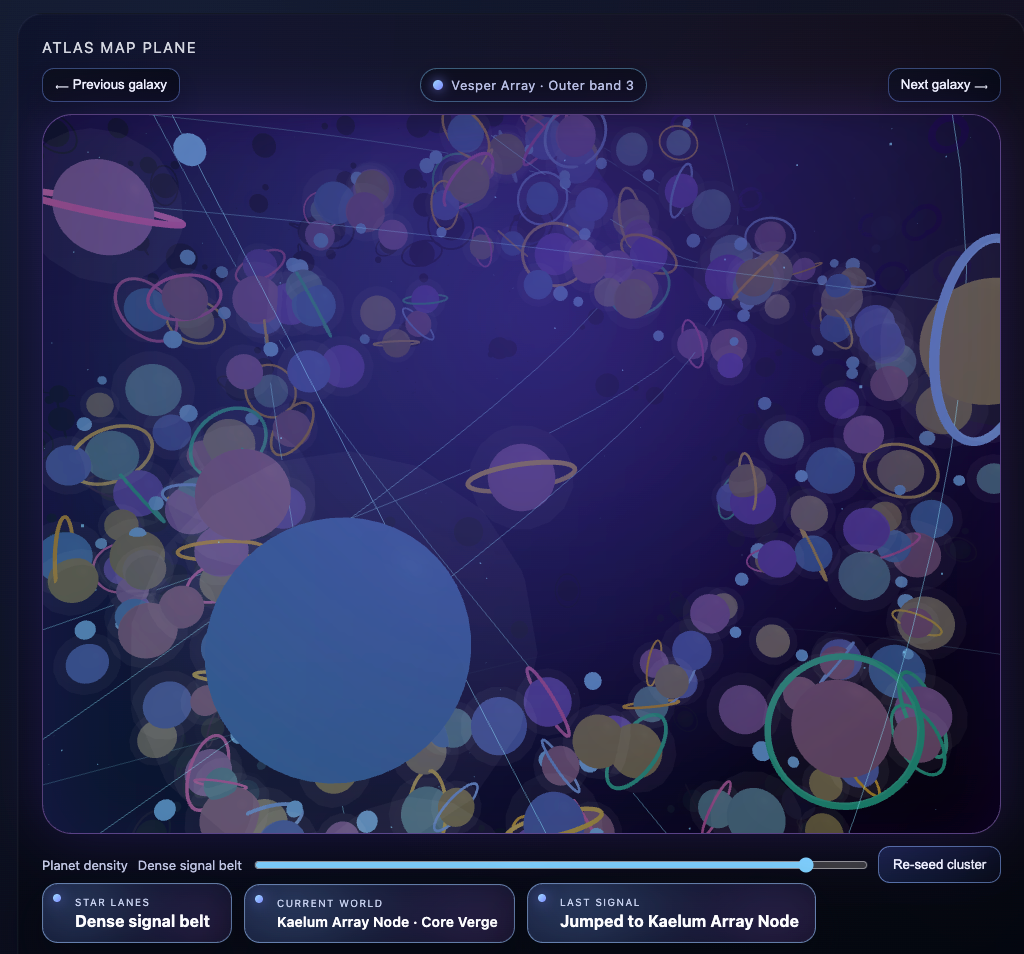}
  \caption{Atlas map plane close-up. The user adjusts the planet density slider to reshape the visible cluster (dense signal belt) while keeping navigation stable via galaxy controls and persistent node identifiers.}
  \label{fig:sci-fi-demo2-density}
\end{figure}

\begin{figure}[ht!]
  \centering
  \includegraphics[width=0.6\linewidth]{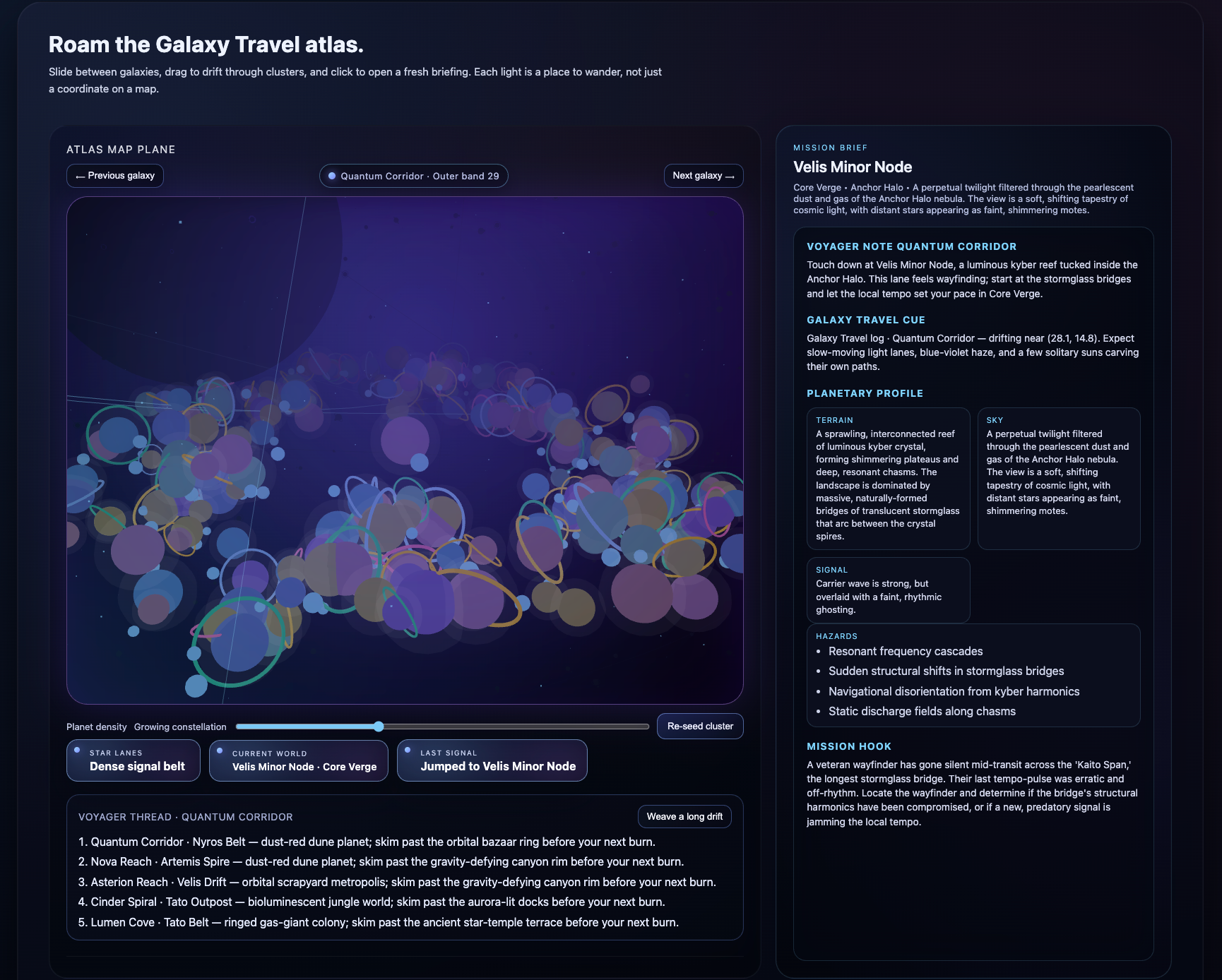}
  \caption{Click-to-generate interaction. Selecting \emph{Velis Minor Node} triggers the agent to produce a structured mission brief (profile cards for terrain/sky/signal/hazards) plus a narrative mission hook, while the voyager thread summarizes a multi-stop route.}
  \label{fig:sci-fi-demo3-velis}
\end{figure}

\begin{figure}[ht!]
  \centering
  \includegraphics[width=0.6\linewidth]{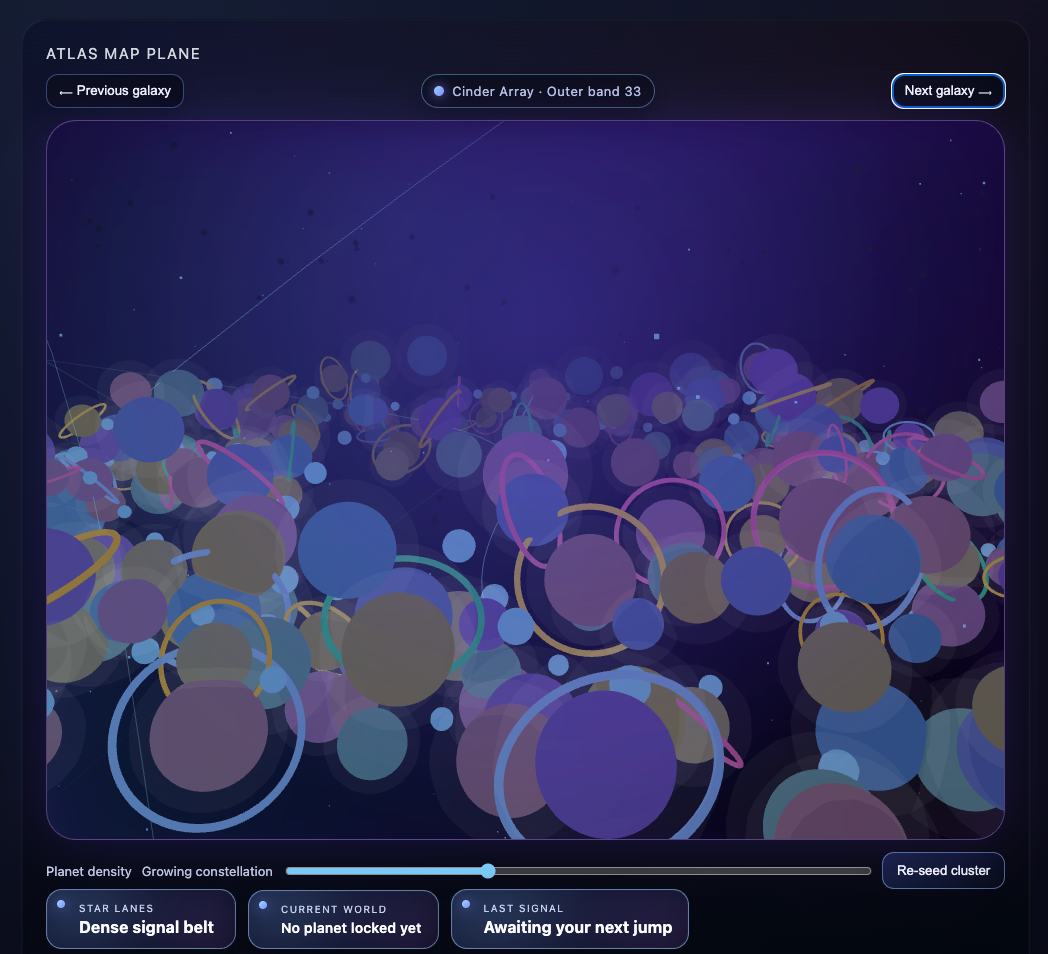}
  \caption{Switching to a new galaxy view. The user steps to another procedurally generated region (\emph{Cinder Array}) and receives a fresh, dense, clickable layout before selecting any specific world.}
  \label{fig:sci-fi-demo4-newgalaxy}
\end{figure}

\begin{figure}[ht!]
  \centering
  \includegraphics[width=0.6\linewidth]{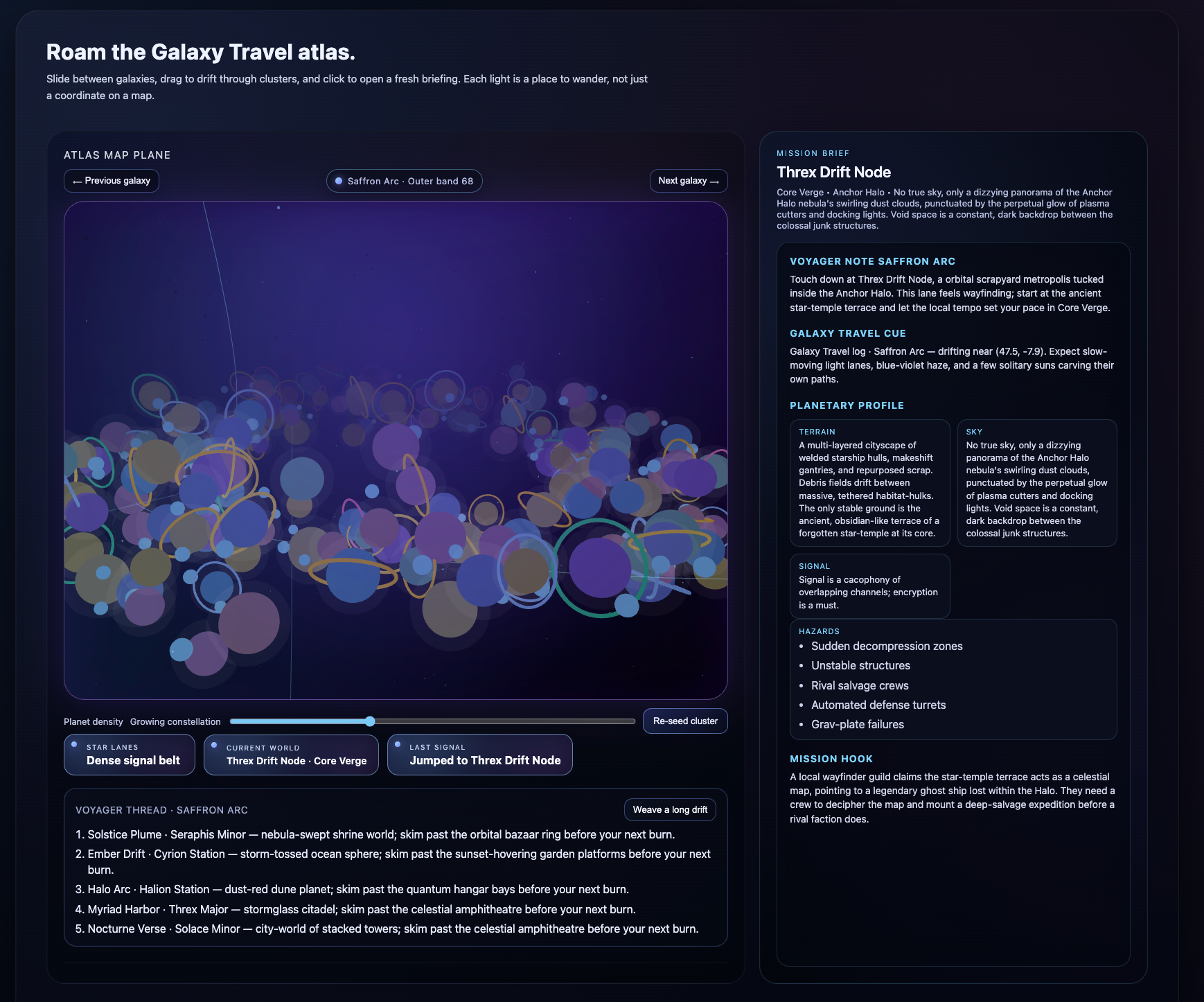}
  \caption{Theme and content variation across worlds. Choosing \emph{Threx Drift Node} yields a contrasting setting (scrapyard-metropolis flavor) with a different hazard list and signal description, while the UI structure remains the same.}
  \label{fig:sci-fi-demo5-threx}
\end{figure}

\begin{figure}[ht!]
  \centering
  \includegraphics[width=0.6\linewidth]{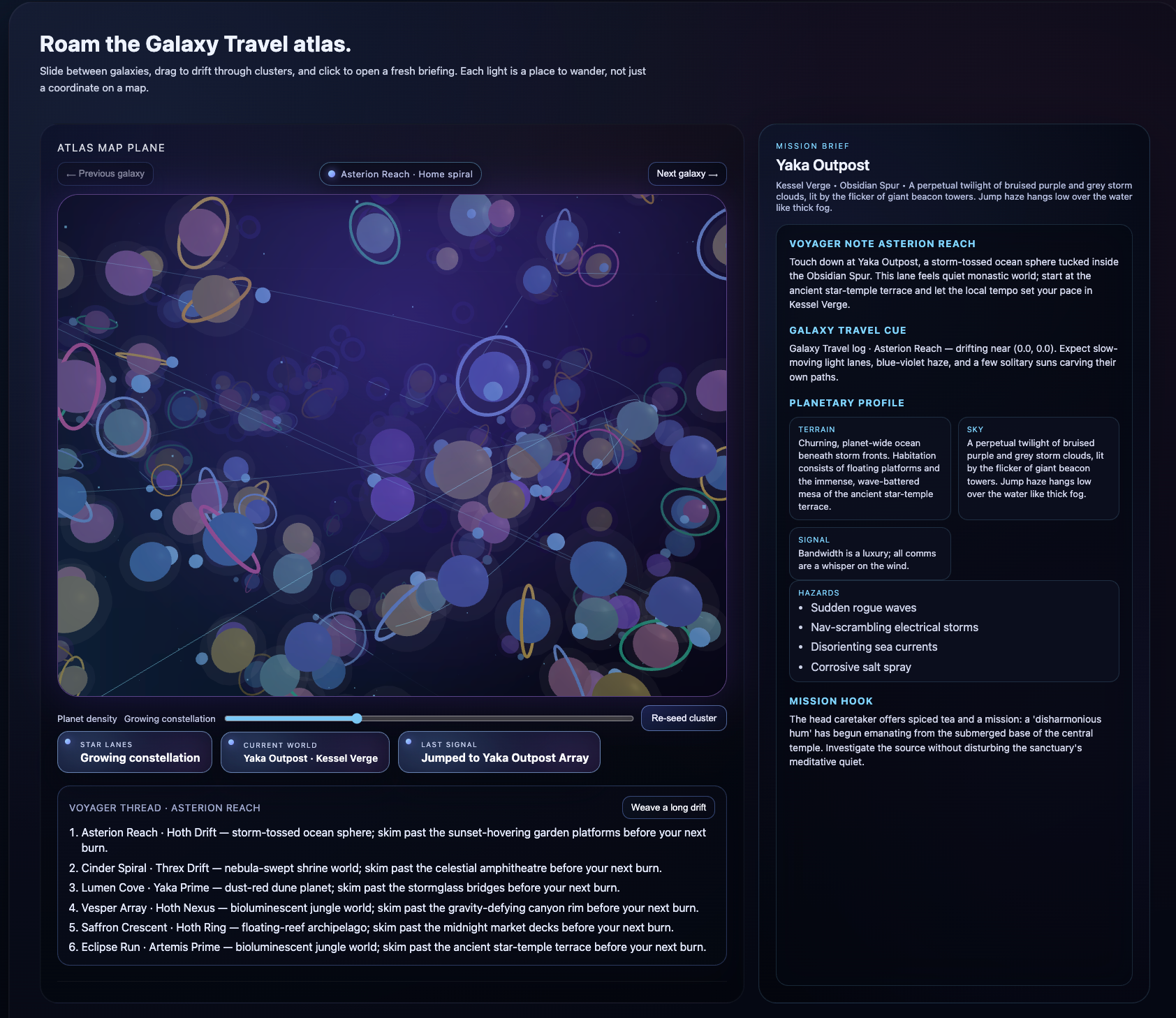}
  \caption{Another node selection with a different “vibe.” \emph{Yaka Outpost} renders as an oceanic outpost scenario with its own terrain/sky/signal/hazards cards and a new mission hook, illustrating location-conditioned generation.}
  \label{fig:sci-fi-demo6-yaka}
\end{figure}

\begin{figure}[ht!]
  \centering
  \includegraphics[width=0.6\linewidth]{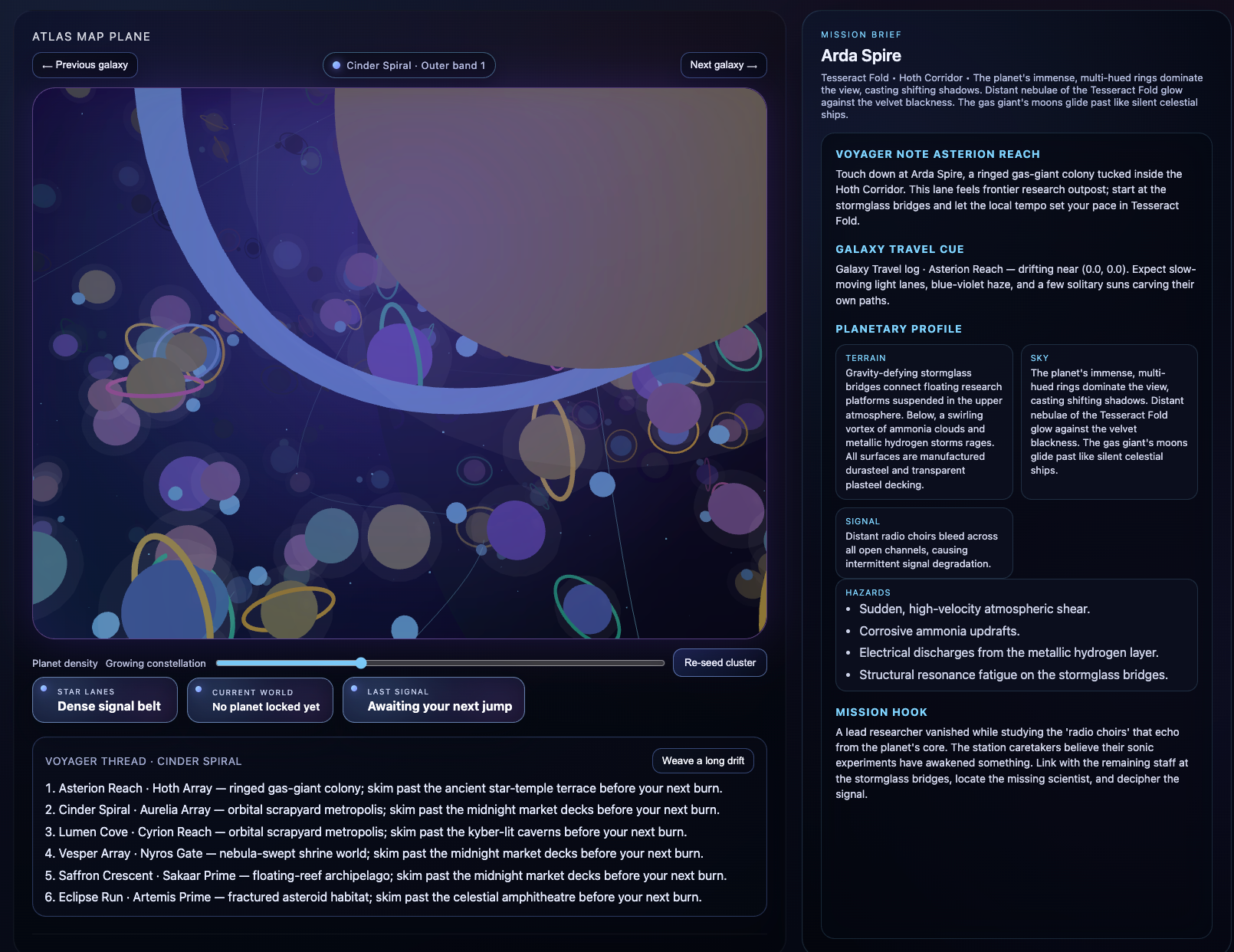}
  \caption{Ringed gas-giant colony example. \emph{Arda Spire} demonstrates how the same atlas interaction pattern supports visually and narratively distinct worlds (e.g., ring structures, atmospheric hazards, and a different exploration prompt).}
  \label{fig:sci-fi-demo7-arda}
\end{figure}

\begin{figure}[ht!]
  \centering
  \includegraphics[width=0.6\linewidth]{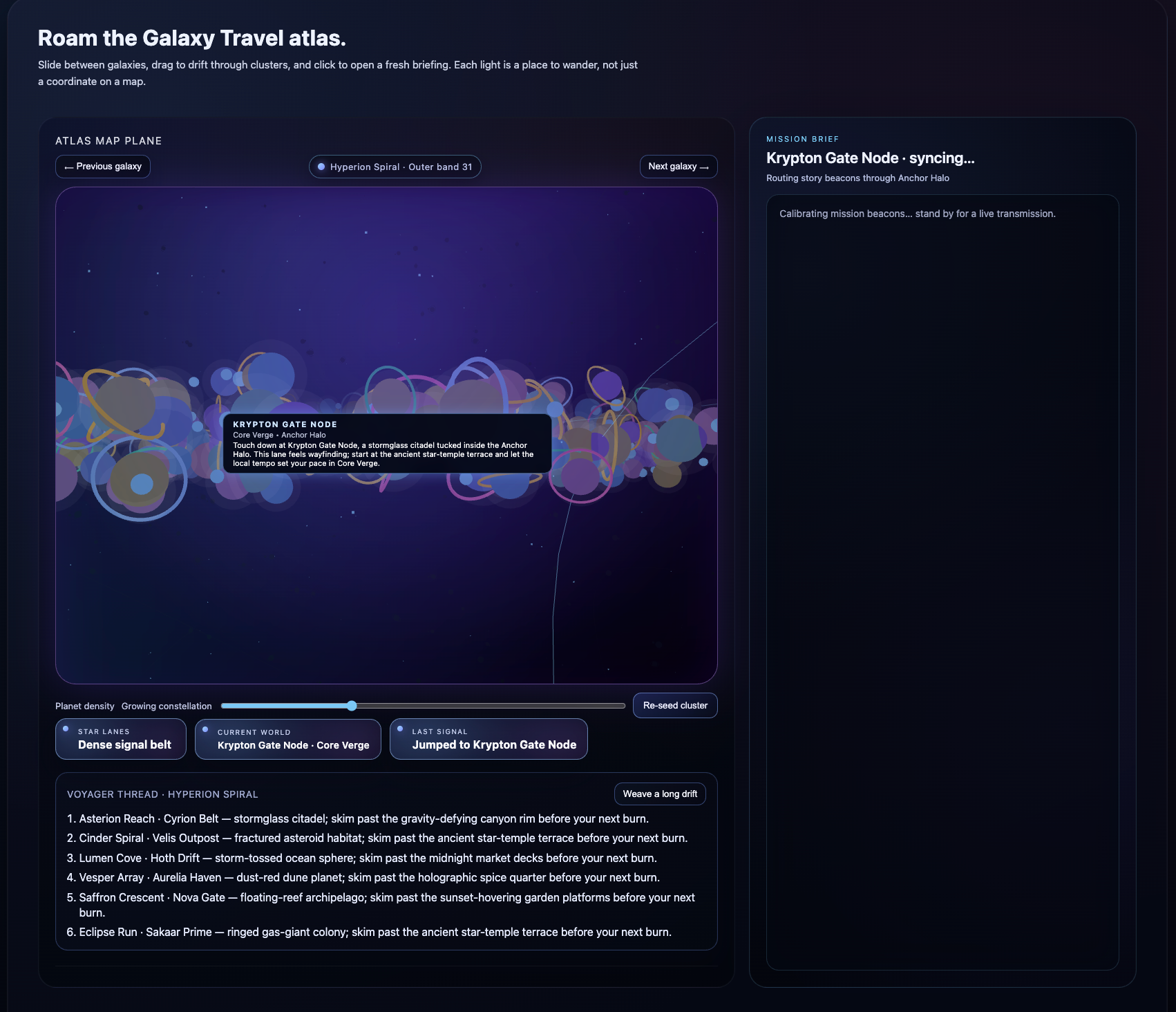}
  \caption{Generation-in-progress (“syncing”) state. After selecting \emph{Krypton Gate Node}, the mission panel enters a live calibration/loading phase while the agent prepares the brief, keeping the interface responsive during content generation.}
  \label{fig:sci-fi-demo8-syncing}
\end{figure}

\begin{figure}[ht!]
  \centering
  \includegraphics[width=0.6\linewidth]{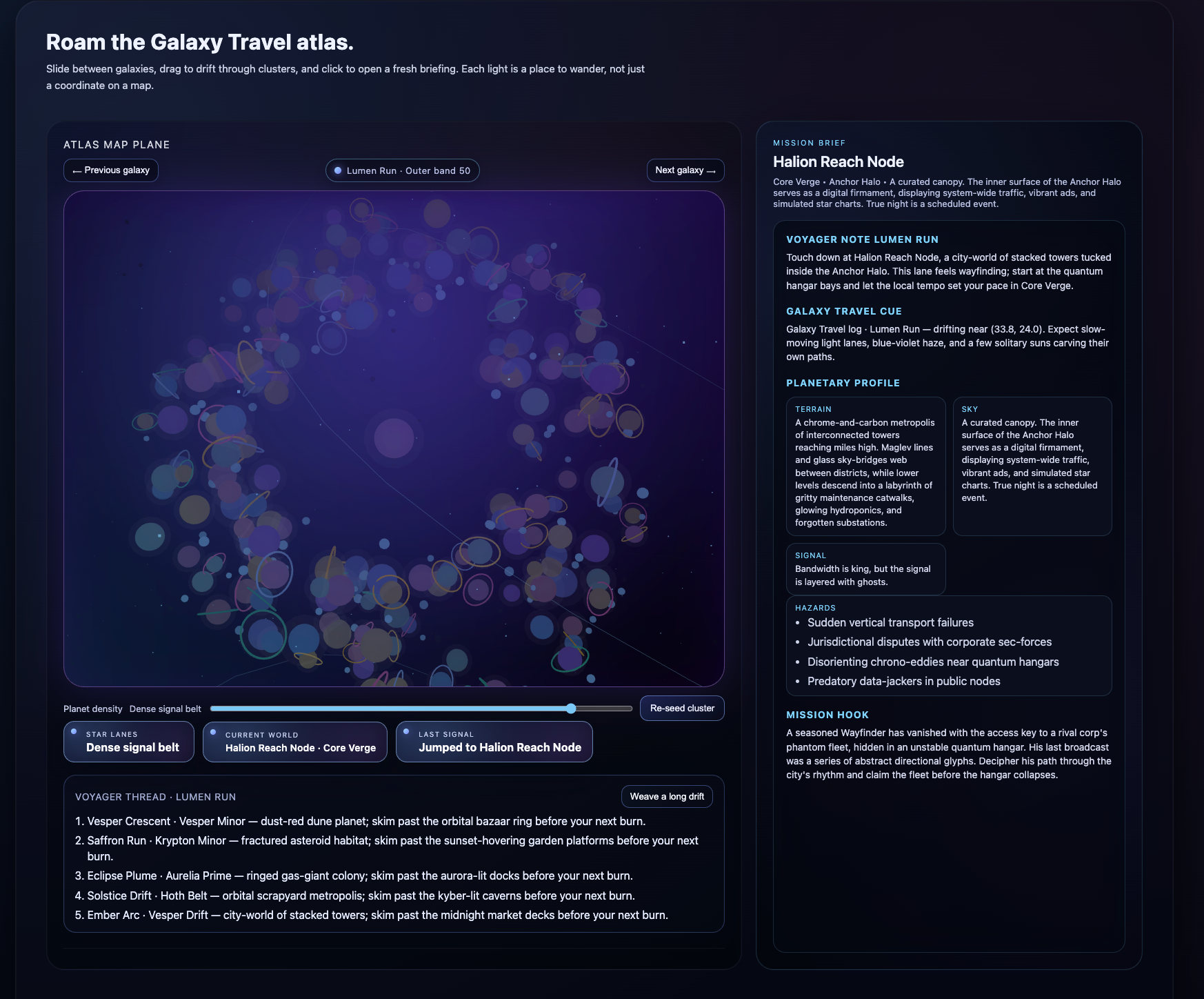}
  \caption{City-world mission brief. \emph{Halion Reach Node} illustrates another distinct profile: a dense metropolitan terrain with different hazards and a separate mission hook, showing repeatable structure with high-variance content.}
  \label{fig:sci-fi-demo9-halionreach}
\end{figure}

\begin{figure}[ht!]
  \centering
  \includegraphics[width=0.6\linewidth]{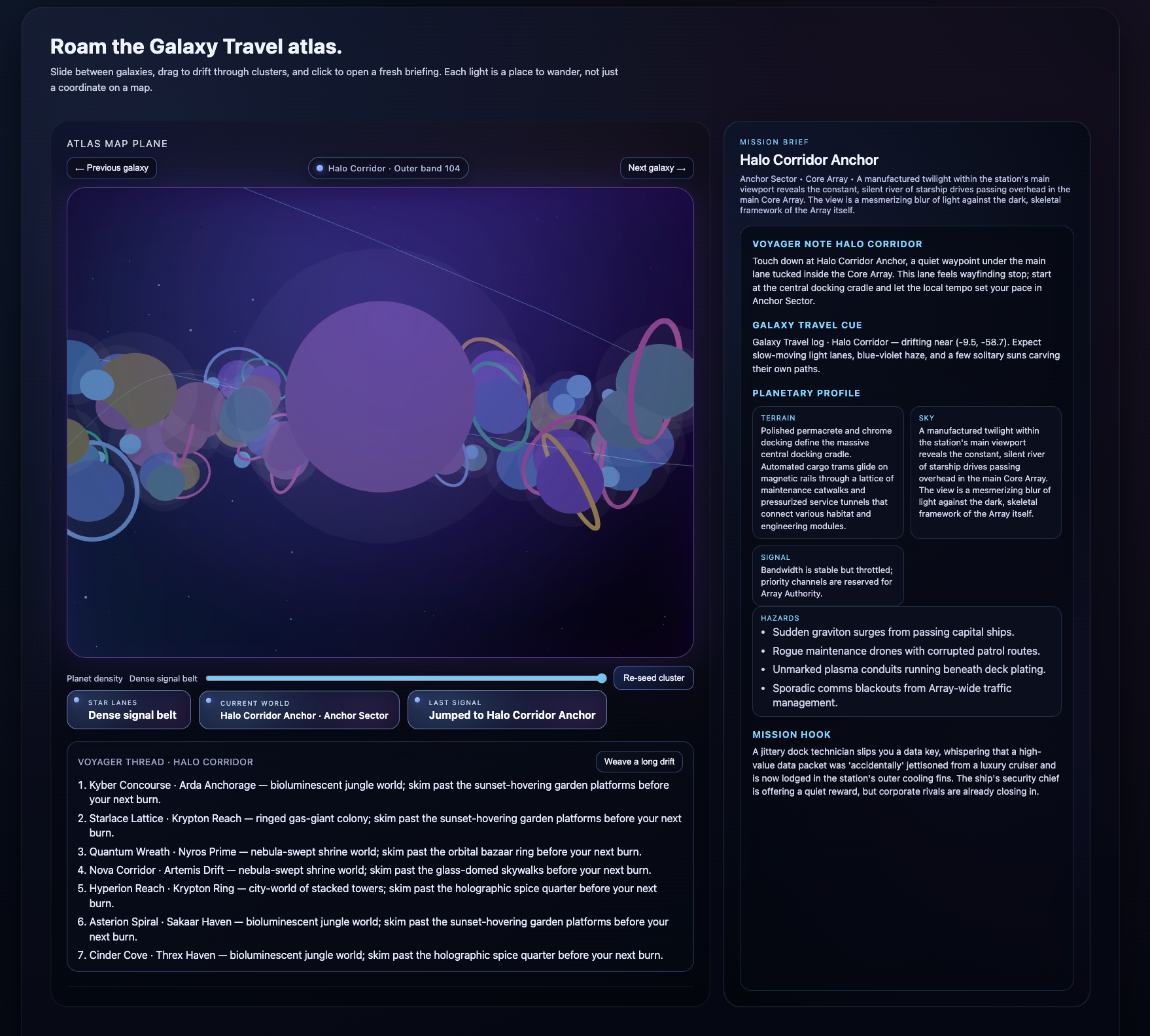}
  \caption{Anchor-type destination. \emph{Halo Corridor Anchor} highlights that the atlas supports non-planet nodes as well, while the Voyager thread continues to stitch worlds into longer exploratory routes across galaxies.}
  \label{fig:sci-fi-demo10-haloanchor}
\end{figure}

\begin{figure}[ht!]
    \centering
    \includegraphics[width=0.6\linewidth]{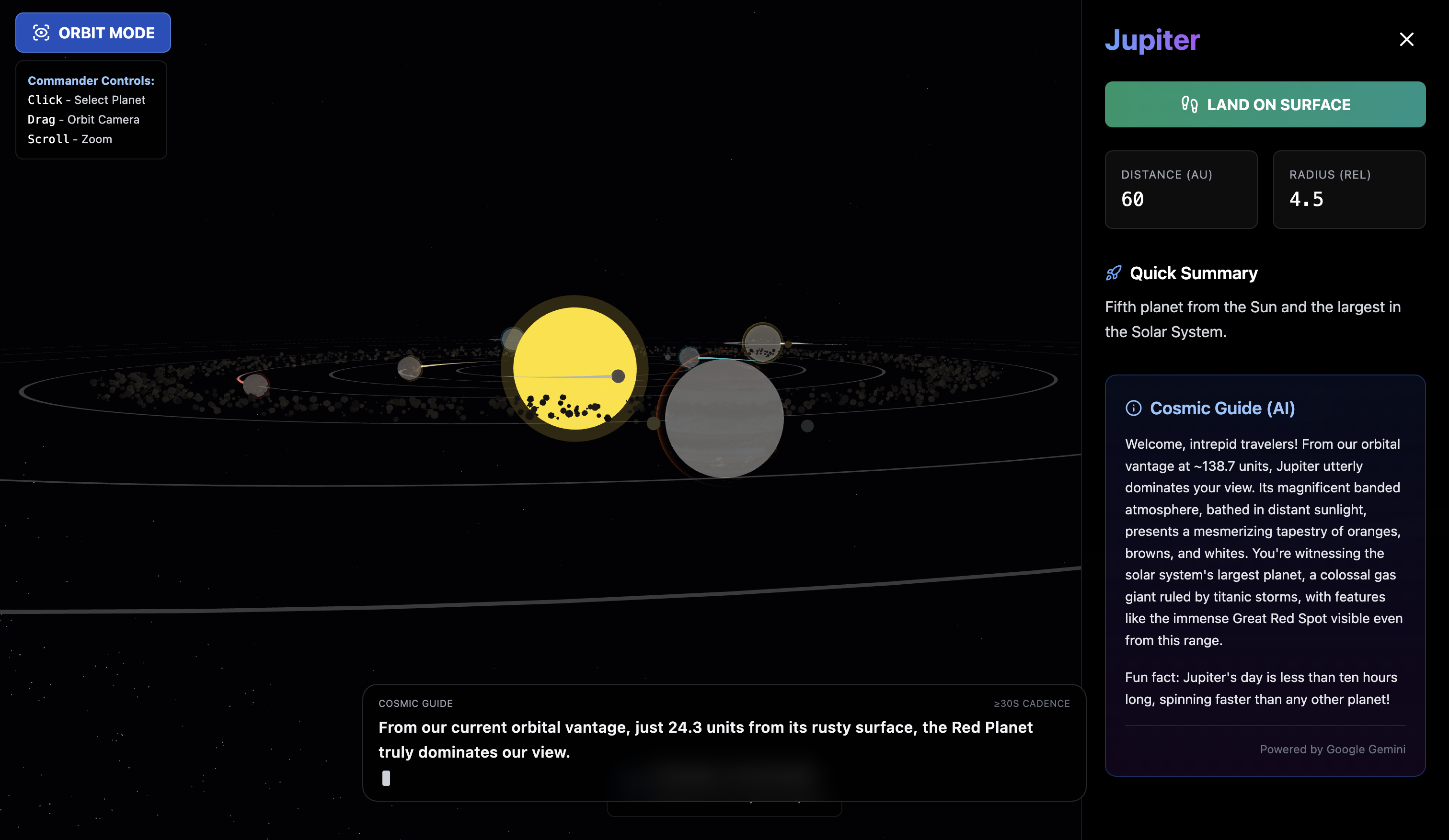}
    \caption{Cosmic Voyager: User can click on the planet to view its description, which is generated by LLM (Gemini 2.5 flash). The AI narration appears at the bottom as a subtitle, showing the description of the user's view.}
    \label{fig:Cosmic Voyager2}
\end{figure}

\begin{figure}[ht!]
    \centering
    \includegraphics[width=0.6\linewidth]{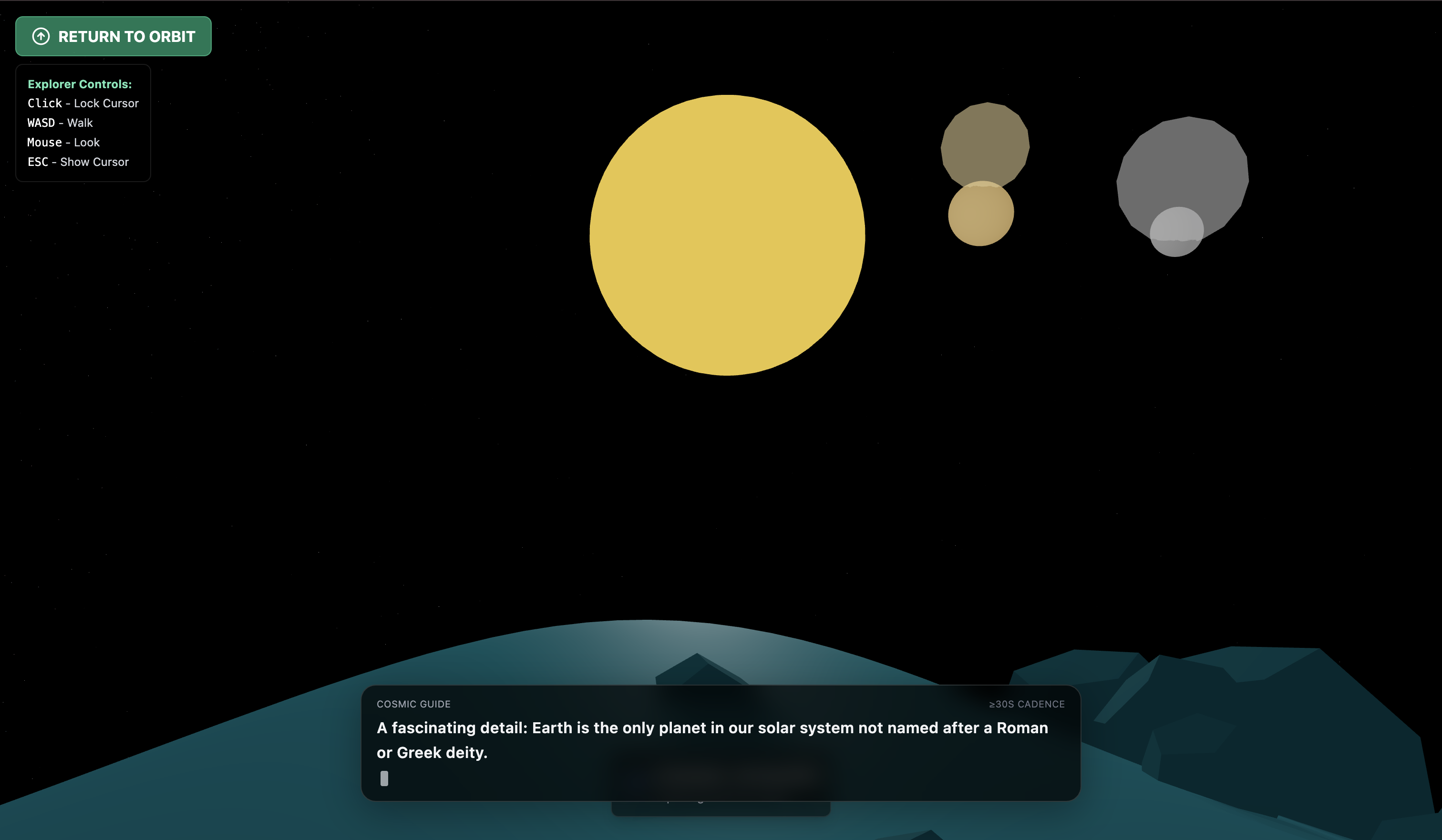}
    \caption{Cosmic Voyager: The demo allows the user to land on the planet and walk on it. The terrain is generated by an LLM with rocks and small mountains on it. The day/night light effect also cast shadow on the rocks. The AI narration persists in this mode.}
    \label{fig:Cosmic Voyager3}
\end{figure}

\begin{figure}[ht!]
    \centering
    \includegraphics[width=0.6\linewidth]{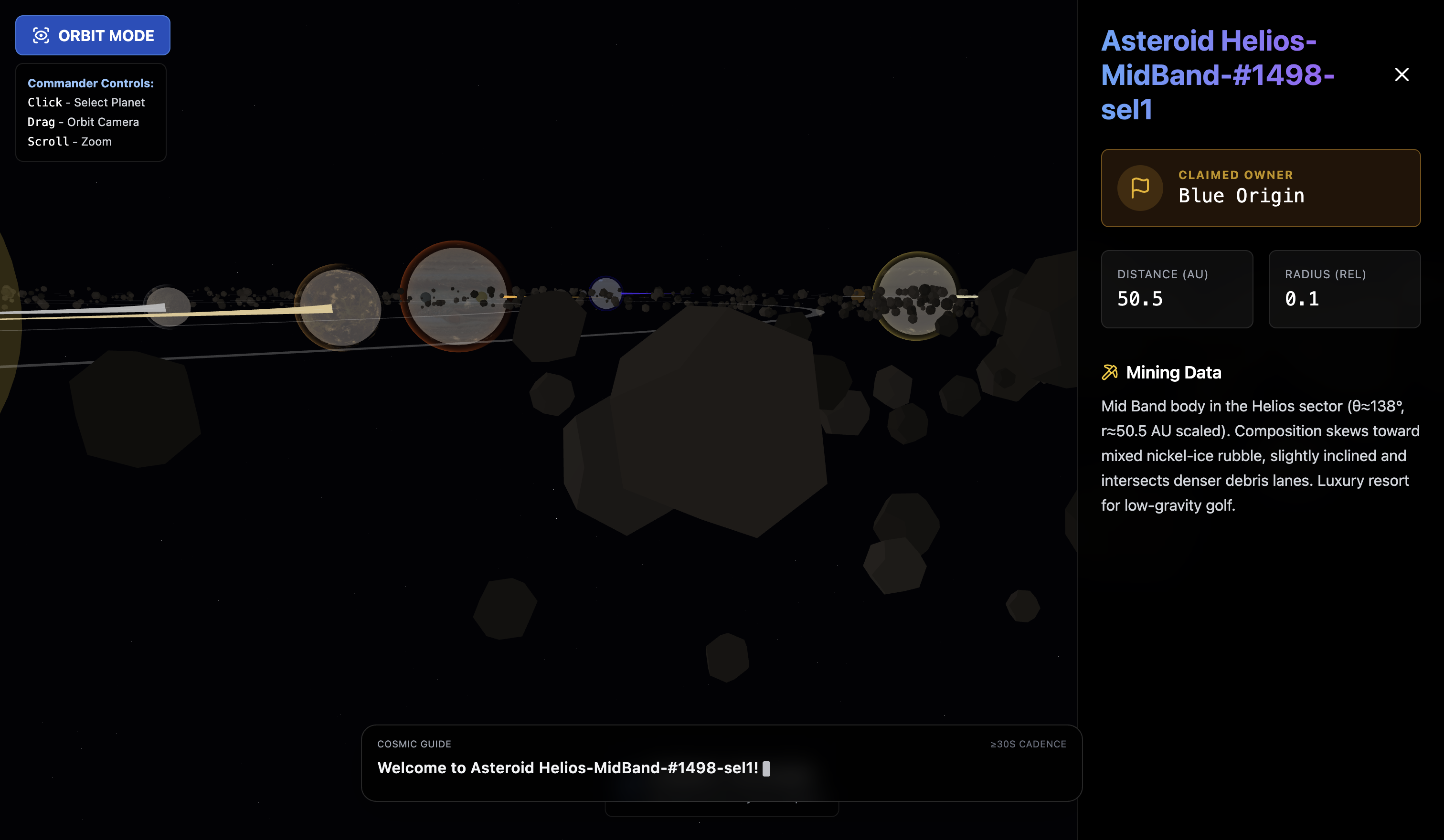}
    \caption{Cosmic Voyager: User can click on an asteroid belt and see its description. The description includes the owner, the mining data, and the size, which is based on the position.}
    \label{fig:Cosmic Voyager4}
\end{figure}

\end{document}